\documentclass[11pt]{article}

\usepackage[preprint]{acl}

\usepackage{times}
\usepackage{latexsym}

\usepackage[T1]{fontenc}

\usepackage[utf8]{inputenc}

\usepackage{microtype}

\usepackage{inconsolata}

\usepackage{graphicx}
\usepackage{hyperref}

\usepackage{dirtytalk}
\usepackage{enumitem}
\usepackage{multirow}
\usepackage{booktabs}
\usepackage[dvipsnames]{xcolor}
\usepackage{amsmath}
\usepackage{tcolorbox}
\usepackage{float}

\usepackage{etoolbox}
\patchcmd{\quote}{\rightmargin}{\topsep 0.3em \partopsep 0pt \leftmargin 0.1em \rightmargin}{}{}

\title{Aligning Language Models for Differentially Private Text Generation:\\A Human-centered Pathway for Better Acceptability and Usability}

\author{
 \textbf{Stephen Meisenbacher\textsuperscript{1,2,3,4}}{\normalfont ,}
 \textbf{Evan Christopher\textsuperscript{1}}{\normalfont ,}
 \\
 \textbf{Jana Diesner\textsuperscript{1,2,3,4}}, \and 
 \textbf{Florian Matthes\textsuperscript{1,4}}
\\
 \textsuperscript{1}Technical University of Munich, School of Computation, Information and Technology\\
 \textsuperscript{2}Technical University of Munich, School of Social Sciences and Technology\\
 \textsuperscript{3}Munich Center for Machine Learning, Munich, Germany\\
 \textsuperscript{4}Munich Data Science Institute, Munich, Germany\\
 \small{
   \textbf{Correspondence:} \href{mailto:stephen.meisenbacher@tum.de}{stephen.meisenbacher@tum.de}
 }
}

\begin{document}
\maketitle
\begin{abstract}
Integrating Differential Privacy (DP) into text privatization has been bolstered by leveraging generative models for private text generation. However, the inherent requirement of DP for noise injection presents problems for coherent text outputs. 
Grounded in the theoretical notion of \textit{acceptability}, our work confirms, through a survey with 294 humans and a subsequent thematic analysis, that DP text generation is generally not human acceptable. Seeking to remedy this, we use the human preferences collected from our survey study to align the decoder of \textsc{DP-BART}, a seminal method in document-level, encoder-decoder DP text generation. We show that alignment with human preferences improves coherence and acceptability, without loss to (and sometimes gaining) utility and empirical privacy. Based on our findings, we argue that addressing the usability challenge in DP text privatization must be grounded in human preferences, which can in turn inform and improve the practicality of technical solutions.
\end{abstract}

\section{Introduction}
The promise of applying Differential Privacy (DP) \cite{dwork2006differential} to text privatization techniques primarily rests in the ability to do so with formal privacy guarantees, as opposed to ad hoc methods lacking such guarantees \cite{near2025guidelines}. While providing such guarantees makes differentially private text privatization a viable option for protections against linkage and inference attacks \cite{meisenbacher-matthes-2024-thinking}, recent work has shed light on considerable issues regarding usability, output coherence, and human acceptance \cite{klymenko-etal-2022-differential,igamberdiev-habernal-2023-dp,meisenbacher-etal-2025-investigating,arnold-2026-differentially}.

The root of the issue can be attributed to the necessitation of DP for \textit{noise injection}, where some form of randomization is built into DP mechanisms to achieve required indistinguishability levels under a chosen \textit{privacy budget} ($\varepsilon$) \cite{fernandes2019generalised}. This fact, which is the grounds for the formal privacy guarantees, seems to be at odds with producing coherent and usable, but private output texts, and such \say{noisification} is at the center of the well-studied \textit{privacy-utility trade-off} \cite{meisenbacher-etal-2025-impact}. The tangible consequences of stricter DP, therefore, often is \say{noisy} text that is lacking natural flow, grammatically incorrect, or generally incoherent \cite{mattern-etal-2022-limits}.

In a recent study, \citet{meisenbacher-etal-2025-investigating} find that current DP text privatization faces significant usability issues, where they conclude that human perceptions will be negative if text outputs are not coherent and \textit{acceptable}. We ground our work in acceptability, which has foundations in theoretical linguistics \cite{chomsky1965aspects} and more recently, in NLP evaluation and language model alignment \cite{warstadt2019neural}. We extend this notion to DP text privatization as a criterion for estimating human acceptance, which must be weighed against privacy protections in addition to \textit{utility}.

To operationalize acceptability in DP text privatization, we investigate the usefulness of language model alignment based on human-sourced preferences for acceptable privatization outputs. This is rooted in the previously ignored point that the pre-trained models used as the underlying component for (generative) DP text privatization are not equipped to handle noise. We posit that alignment can rectify this, wherein text generation from noisy latent representations can be steered towards more coherent and acceptable outputs. As a starting point, we experiment with the \textsc{DP-BART} method \cite{igamberdiev-habernal-2023-dp}, which utilizes a pre-trained \textsc{BART} model for DP text generation. 

As an initial finding, we learn that a composite metric that captures semantic fidelity, fluency, and readability can serve as a representative measure of \textit{acceptability} for privatized texts. This is confirmed through a preference survey study with 294 humans from around the globe. We use the collected preferences to perform DPO alignment of the \textsc{DP-BART} decoder component, tested on nine training configurations. The evaluation of the aligned models reveals that privatization preference alignment not only improves the acceptability of privatized texts, but also maintains or strengthens utility, empirical privacy, and their trade-offs. We conclude that human-grounded alignment training is vital for language model-based DP text privatization.

With our work, we make the following contributions to the field of DP text privatization, particularly in promoting (human-grounded) usability:

\begin{enumerate}
    \itemsep -0.4em
    \item We formalize and operationalize \textit{acceptability} for DP texts, grounded in a scoping literature review of 103 works, enabling the creation of a text privatization preference dataset.
    \item We conduct a survey study with 294 human participants, which validates our selected acceptability metrics and surfaces five major themes for acceptable DP text privatization.
    \item We investigate the feasibility of alignment training for improving the acceptability of DP text privatization, performed through a case study with the \textsc{DP-BART} architecture.
    \item We share a full replication package, including preference dataset, survey, training pipeline, and aligned models: \url{https://github.com/crustypluto19/aligning-DP}
\end{enumerate}

\section{Foundations and Related Work}
\paragraph{Differential Privacy.} DP \cite{dwork2006differential} is a theoretical framework for quantifying privacy protection in data sharing. DP bounds how much any single individual's contribution to a dataset (e.g., a row in a table) can influence the impact of a computation on that data, formalized by:
\[
    Pr[\mathcal{M}(D) \in S] \leq e^{\varepsilon} Pr[\mathcal{M}(D') \in S]
\]

Intuitively, this inequality enforces, for any two \textit{adjacent} datasets $D$ and $D'$ (differing in one individual) and for any output in the space $S$, that such an output cannot be attributed to $D$ or $D'$, within some probability bound. This bound is governed by the privacy parameter (budget) $\varepsilon$, and the \textit{privacy loss} is represented by $e^\varepsilon$. In this, the amount of \textit{noise} needed to make outputs \textit{indistinguishable} is controlled by the $\varepsilon$ input of a DP mechanism $\mathcal{M}$. 

\paragraph{DP Text Privatization.}
The integration of DP into the unstructured domain of text and NLP has received growing attention in the last decade, largely due to advancements in application areas such as Machine Learning \cite{10.1145/2976749.2978318} or NLP \cite{ICLR2024_8cb564df,ramesh-etal-2024-evaluating}. While the application of DP within NLP is quite diverse \cite{hu-etal-2024-differentially}, our work focuses on the task of \textit{DP text privatization}, where an input text is transformed under DP guarantees into a private output, typically achieved with \textit{local} DP:
\vspace{-3pt}
\[
    Pr[\mathcal{M}(x) = z] \leq e^{\varepsilon} Pr[\mathcal{M}(y) = z]
\]
This presents a stricter form of DP, in which \textit{all} input pairs $x$ and $y$ must be indistinguishable, under $\varepsilon$, given a DP output $z$. The privatization unit (i.e, $x$ and $y$) has been interpreted in diverse ways in the context of text privatization, ranging from word- and token-level approaches \cite{feyisetan_balle_2020,yue-etal-2021-differential,meisenbacher-etal-2024-dp,10.1145/3746252.3760888} to full sentence- or document-level rewriting approaches \cite{10.1145/3485447.3512232,mattern-etal-2022-limits,igamberdiev-habernal-2023-dp,meisenbacher-etal-2025-impact}.

A key persistent challenge in the pursuit of usable DP text privatization has been the balancing of formal guarantees with coherent outputs. Critiquing context-unaware word-level approaches, \citet{mattern-etal-2022-limits} makes a case for generative approaches, where other works \cite{arnold-etal-2023-driving,arnold-etal-2023-guiding} aim to rectify these inherent shortcomings. Generative approaches, though, face their own unique challenges, predominantly with inflated privacy budgets \cite{igamberdiev-habernal-2023-dp} and non-coherent outputs in stricter privacy regimes \cite{meisenbacher-etal-2025-impact}. Regardless of privatization unit or underlying mechanism, non-coherent or nonsensical outputs will deter human acceptance of DP text privatization \cite{meisenbacher-etal-2025-investigating}, a challenge we address in this work.

\paragraph{Acceptability in linguistics and NLP.}
Research spanning cognitive science to NLP has investigated the notion of \textit{acceptability}, contributing to the theoretical \cite{sprouse2013acceptability,lau} and technical understanding of acceptability in language \cite{lau-etal-2015-unsupervised,warstadt2019neural}.
Particularly in the NLP research community, numerous factors affecting acceptability judgments have been explored, including multilingual settings \cite{zhang-etal-2024-mela}, the impact of context \cite{sinha-etal-2023-language}, and general inquiries into the feasibility of LLM-based assessments \cite{golan2023testing}. While acceptability is often applied in the context of evaluating generative outputs, it has not been formalized in the context of \textit{private} text generation.  

\section{\textit{Acceptability} of Privatized Text}
\label{sec:accept}
To build a definition of what \textit{acceptability} may entail in the context of DP text privatization, we conducted a scoping literature review to survey existing metrics. We opted to compile a composite metric, following the notion that evaluation in text privatization is heterogeneous \cite{ren-etal-2025-measure}. A scoping review is appropriate because we sought to map the breadth of evaluation metrics across diverse domains rather than to synthesize evidence for a specific intervention \cite{munn2018systematic}.

\paragraph{Search strategy.}
Formulating the search strategy began with the selection of three key desiderata for metrics in the context of DP text privatization:
\begin{enumerate}
    \itemsep -0.4em
    \item \textit{Semantic fidelity}: how well the generated text preserves meaning for downstream NLP tasks.
    \item \textit{Fluency}: how closely the generated text aligns with what humans expect of natural language.
    \item \textit{Readability}: whether the text remains interpretable and readable under privatization.
\end{enumerate}

The selection of these axes was guided by our own experience in the field, as well as prior surveys in the research community \cite{klymenko-etal-2022-differential,hu-etal-2024-differentially,meisenbacher-etal-2025-investigating}.

From this basis, we created one primary search string (S1) and one supplementary string (S2):

\begin{itemize}
    \scriptsize
    \itemsep -0.4em
    \item[S1.] \textit{``differential privacy'' AND (``natural language processing'' OR ``nlp'') AND ``text generation'' AND ``evaluation metric'' AND (``semantic'' OR ``coherence'' OR ``utility'' OR ``representation'')}
    \item[S2.] \textit{``metric'' AND ``readability evaluation'' AND (``natural language processing'' OR ``nlp'')}
\end{itemize}

While S1 was primarily motivated by our established desiderata, S2 was included to cast a wider net to metrics that we (or the previous) literature had not considered in the context of DP text privatization evaluation. These two search strings were used in three popular academic databases (Google Scholar, Scopus, ACM Digital Library).

Finally, to supplement the structured search, we augmented our scoping literature review with a snowball strategy from two seed surveys \cite{celikyilmaz2021evaluationtextgenerationsurvey,10.1145/3485766}, opportunistically and recursively searching for references from these seed works. For all searches, we limited our review to works published from 2024 onwards.

\paragraph{Study selection.}
S1 yielded 313 total initial results, with 24 duplicates. S2 was stopped after the first 50 results for each database due to substantial overlap with S1. With these, a two-phase screening was applied, where the titles and abstracts of all candidates were screened for general relevance to our search. Papers passing this stage (103 in total) were full-text screened, where our team developed inclusion criteria and exclusion criteria (Appendix \ref{sec:litreview}) to guide the final selection. The screening process was carried out by one primary researcher, and the final selected papers were validated (via title and abstract) by the remainder of the team.

\paragraph{Metric selection.}
From the final set of selected papers, the objective became to survey the collection of evaluation metrics contained therein, namely those aimed at evaluating text quality, coherence, fidelity, or in sum, \textit{acceptability}. 

Our team extracted each metric via the full-text screens and coded them along three axes: (1) \textit{Facet}: assignment to one of our three desiderata, (2) \textit{Granularity}: on which unit the metric operates, and (3) \textit{Type}: reference-free or reference-based.

The full collection of extracted metrics is found in Table \ref{tab:metrics-scoping-review} of Appendix \ref{sec:litreview}. From this full set, we proceeded to select a representative subset that (1) represents all three defined desiderata, (2) combines token- and document-level scores (but not corpus-level), and (3) includes both reference-free and reference-based measures. This final set of metrics is presented in Table \ref{tab:chosen-metrics}. From this, we operationalize \textit{acceptability} for DP text privatization:

\begin{quote}
    \small
    \textit{Acceptability: a human-centered, continuous measure of how coherent, readable, and semantically faithful a private text is.} 
\end{quote}

All metrics are normalized, and each facet score represents the unweighted average of both component metrics. The precise formulations of these metrics and scores are found in Appendix \ref{sec:metrics}.

\begin{table}[t]
\centering
\small
\resizebox{\linewidth}{!}{
\begin{tabular}{c|c|c|c}
\toprule
\textbf{Metric} & \textbf{Facet} & \textbf{Granularity} & \textbf{Type} \\
\midrule
BERTScore \cite{zhang2020bertscore} & \multirow{2}{*}{Semantic Fidelity} & Token/Doc & \multirow{2}{*}{Ref} \\
\cline{1-1} \cline{3-3}
BLEURT \cite{sellam-etal-2020-bleurt}  &       & Document   &     \\
\hline
PPL \cite{jelinek1977perplexity} & \multirow{2}{*}{Fluency}           & Token & \multirow{2}{*}{Free}  \\
\cline{1-1} \cline{3-3}
CoLA \cite{warstadt2019neural}  &             & Document   &      \\
\hline
CLI  \cite{coleman1975computer}     & \multirow{2}{*}{Readability}       & Document   & \multirow{2}{*}{Free} \\
\cline{1-1} \cline{3-3}
Dale-Chall \cite{dale1948formula} &                                    & Document   & \\
\bottomrule
\end{tabular}
}
\caption{Metrics for \textit{acceptability}, based on our scoping literature review and metric selection.}
\label{tab:chosen-metrics}
\end{table}

\section{Surveying Privatization Preferences}
With the evaluation of acceptability as a motivation, we first construct a DP privatization preference dataset, which then serves as the basis for an online survey with 294 human participants. This survey provides the foundation for a validation of our acceptability measures, as well as seeds the case study we describe in the subsequent section.

\subsection{Constructing a Preference Dataset}
\label{sec:construct}
The goal of the preference dataset is to span a meaningful and representative range of DP text privatization outputs, drawing from diverse text sources and covering various privacy regimes ($\varepsilon$ budgets). We describe the process followed to build this dataset.

\paragraph{Methods.}
\hspace{-10pt} We use three text privatization methods: (1) \textbf{\textsc{DP-BART}} \cite{igamberdiev-habernal-2023-dp}, an encoder-decoder model that privatizes text by injecting noise into the latent representation, (2) \textbf{\textsc{DP-Prompt}} \cite{utpala-etal-2023-locally}, a prompt-based method that applies DP temperature sampling during decoding, and (3) \textbf{\textsc{DP-MLM}} \cite{meisenbacher-etal-2024-dp}, a masked LM approach that replaces tokens using the exponential mechanism. 

\paragraph{Datasets.}
To form a diverse basis for the preference dataset and survey, we sample texts from 10 publicly available datasets that feature human-written texts in potentially sensitive areas, including texts in the medical, legal, social media, and review domains. The 10 datasets are introduced more formally in Appendix \ref{sec:datasets}, and  the filtered datasets are included in our repository. From each dataset, we selected at random (seed 67) 25 texts that contained between 50-100 words (for readability in the survey). From these, 12 final texts were chosen manually by our team, guided by two primary criteria: (1) texts that discuss a self-contained topic intelligible without prior context, and (2) texts that contain the fewest errors, typos, or otherwise noisy content. This process was done opportunistically until 12 texts were found per dataset. The final set of 120 texts (also in our repository) thus formed the reference set for all ensuing experiments. 

\paragraph{Privatization calibration.}
With the 120 texts and three privatization methods, the objective became to construct comparative samples, where a base text was privatized under several $\varepsilon$ budgets. This process was necessary to create private outputs of varying quality ranges (with the assumption that lower $\varepsilon$ leads to worse quality), such that clear preferences between outputs could be captured.

The choice of $\varepsilon$ values per privatization mechanism required a calibration process. We opted for three values, i.e., to produce three outputs for each of the 120 texts. We justified this choice in easing the comparative selection task of the survey. Following \citet{meisenbacher-etal-2025-investigating}, we calibrated three \textit{privacy levels} based on the average cosine similarity of privatization outputs with respect to inputs. Each of the 120 selected texts was privatized 20 times per $\varepsilon$ across a sweep (Appendix \ref{sec:calibrate}).

The results of these sweeps and our selection rationale are provided in Appendix \ref{sec:calibrate}. We chose three $\varepsilon$ per mechanism: \textsc{DP-MLM} ($\varepsilon \in \{25, 50, 100\}$), \textsc{DP-Prompt} ($\varepsilon \in \{50, 60, 75\}$), and \textsc{DP-BART} ($\varepsilon \in \{750, 1500, 2500\}$). With the values, we ensure an empirically comparable inclusion of privatized outputs for fundamentally different mechanisms. As such, we make no claims of theoretical comparability, as this is not necessary for and outside the scope of our work.

\paragraph{Candidate scoring and selection.}
We then proceeded to construct the preference dataset, where each of the 120 texts would be represented in nine setups: three mechanisms $\times$ three privacy levels. Within each setup, our objective was to select three representative texts spanning different levels of acceptability, calculated according to our chosen metrics (Section \ref{sec:accept}, Appendix \ref{sec:metrics}). This process is rooted in the fact that given a fixed mechanism and privacy budget, different runs of the exact privatization setup can produce significantly different outputs in terms of acceptability, due largely to random noise injection. With this, the three options per setup form the basis for a single survey question, in which a survey participant was to select which of the three private outputs is most acceptable.

Per mechanism and $\varepsilon$ level, we privatized each of the 120 texts 50 times. Each of the 50 private outputs was rated along our three acceptability scores (semantic fidelity, fluency, readability). From the sorted results (i.e., texts scored by average acceptability score), we performed simple tertile selection, picking the text at the sorted indices of 16, 33, and 49 (33\textsuperscript{rd}, 66\textsuperscript{th}, and 100\textsuperscript{th} percentiles, respectively). In total, we selected $120 \times 9 = 1080$ private text triplets, each comprising a survey question.

\subsection{Surveying Text Privatization Preferences}
The administration of a text privatization preference survey carries a three-fold purpose. Firstly, we validate our choice of acceptability metrics in their alignment with human preferences; we also validate this with LLM-as-a-Judge. Next, the preference responses collected via the survey serve as the foundation for the case study conducted in Section \ref{sec:case}. Finally, the freeform feedback gathered from the survey participants surfaces qualitative criteria users apply when evaluating privatized texts, which we extract in a thematic content analysis.

\definecolor{tumblue}{RGB}{0, 101, 189}
\definecolor{tumlightblue}{RGB}{152, 198, 234}
\begin{figure}[!t]
\small
\begin{tcolorbox}[
  colback=tumlightblue!30,
  colframe=tumblue!80,
  title={\footnotesize Survey Instructions},
  fonttitle=\bfseries\small,
  left=2pt, right=2pt, top=3pt, bottom=3pt
]
\scriptsize
For each question, you will see the original text followed by three
alternative rephrasals. Choose the single alternative you find
\textbf{MOST ACCEPTABLE}: the one that best preserves the meaning
of the original text while being:
\begin{itemize}[noitemsep, topsep=2pt,leftmargin=2em]
    \item \textit{most private} (i.e., least revealing of sensitive
    information),
    \item \textit{easiest to understand} (clearest in topic),
    \item \textit{most coherent and fluent}, and \textit{easiest to read}.
\end{itemize}

\textit{Please use your own understanding of ``privacy''. We do not provide a definition because we want your natural judgment about what feels private.}\\[2pt]
Estimated time: 15 minutes.
\end{tcolorbox}
\vspace{-10pt}
\caption{Introductory survey instructions.}
\label{fig:survey_instructions}
\end{figure}

\paragraph{Survey design.}
We designed a custom survey instrument using Streamlit, which was deployed on the Streamlit Community Cloud. All questions and responses were stored in a MongoDB instance.

The survey opens with an instruction set and contextualization, replicated in Figure \ref{fig:survey_instructions}. Each survey question presents a participant with an original reference text and three privatized alternatives, all produced by a single DP method at a single $\varepsilon$ level. The participant selects the output they find most acceptable, weighing privacy, comprehension, coherence, and readability based on their own understanding of these properties. As a final, optional question, we included a free-form question:

\begin{quote}
    \footnotesize
    \textit{In general, what were your reasoning patterns behind answering the previous questions? How did you decide which texts seemed most acceptable to you? What did you consider when determining whether texts were private or not?}
\end{quote}

We built the survey mechanism to randomly sample questions from our pool of 1080, with at least one text per 10 reference datasets. A balancer was built to ensure all response options were sampled equally over time. Response options were randomized to mitigate ordering bias, but the options (i.e., responses) were realigned in the survey backend.

The survey was piloted with 10 close colleagues and family, which revealed two necessary changes (already reflected in our descriptions above) after a median completion time of 20 minutes: a reduction of the number of reference texts per dataset from the original 15 to 12, and the number of response options from the original 5 (quintiles) to 3 (tertiles). 

\paragraph{Recruitment.}
Survey participants were recruited via Prolific, and responses were collected in progressive batches across multiple geographic regions. We required a minimum age of 18, fluency in English, and membership in the Prolific-vetted \textit{comparative reasoning} group. The target number of participants was derived from three constraints: a target redundancy factor of four responses per question (i.e., a value greater than the number of response options), a survey length of 15 questions per participant targeting a 20-minute completion time (informed by the pilot), and 1080 total questions. Thus, we obtained a minimum required sample of $N = \frac{1080 \cdot 4}{15} = 288$ total participants. We set this number to 300 to provide a buffer. We compensated each completion with a fair wage of $\pounds 3$ ($\pounds 9$/hr). 294 participants successfully completed the survey; their demographics are reported in Appendix \ref{sec:demo}.

\subsection{Survey Analysis}
\label{sec:analysis}

\paragraph{Raw responses and metric alignment.}
As seen in Table \ref{tab:model_tertile_toprank}, survey participants selected the highest-scoring candidate (tertile 3, according to our acceptability score) 48.7\% of the time ($\chi^2$ = 493.78\footnote{\scriptsize Checks whether humans preferred higher tertiles over random chance.}, $p$ < 0.001). Chosen response options scored noticeably higher than those not selected (0.33 vs. 0.06; $U$ = 10,835,826\footnote{\scriptsize Compares the acceptability scores of those selected against not selected.}, $p$ < 0.001, r = 0.108). Inter-annotator agreement yields a Krippendorff's $\alpha$ = 0.21, indicating high variation in what humans consider acceptable privatized text. However, we also attribute this partly to the redundancy factor (4), which limits each survey question to only four raters.

\begin{table}[t]
\small
\centering
\begin{tabular}{lrrr}
\toprule
\textbf{Annotator} & \textbf{T1 (\%)} & \textbf{T2 (\%)} & \textbf{T3 (\%)} \\
\midrule
Human majority  & 22.5 & 28.8 & 48.7 \\
\midrule
GPT-4o mini     & 14.7 & 24.3 & 61.0 \\
Kimi-2.5        & 17.6 & 23.4 & 59.0 \\
DeepSeek-V3.1   & 26.9 & 23.5 & 49.6 \\
Qwen-32B        & 16.9 & 33.8 & 49.3 \\
LLaMA-70B       & 26.2 & 33.1 & 40.6 \\
\bottomrule
\end{tabular}
\caption[Per-annotator tertile preference distribution.]{Per-annotator tertile preference distribution. The highest tertile
(T3) represents top-rank agreement with the acceptability metric's highest-scoring text.}
\label{tab:model_tertile_toprank}
\end{table}

We also test the alignment of the survey responses with five chosen LLM judges (see Table \ref{tab:model_tertile_toprank}). These LLMs were prompted with a mirror image of the survey instructions (Figure \ref{fig:llm_judge_prompt} of the Appendix). We find that the five LLM judges broadly replicate the human preference pattern, with natural individual variations. The top-scoring (acceptable) text is selected 51.9\% of the time on average ($\chi^2$ = 879.94, $p$ < 0.001), and chosen candidates score considerably higher than those not selected (0.37 vs. 0.03; $U$ = 16,745,742, $p$ < 0.001, r = 0.138).

The survey responses and LLM validation support two claims. Firstly, we demonstrate that our chosen composite acceptability score accurately, but not absolutely, mirrors human perceptions of acceptable privatized text outputs. Secondly, these preferences are also ingrained in LLMs, suggesting that they may be leveraged to proxy human perceptions of acceptability in DP text privatization.

\begin{table}[t]
\centering
\small
\resizebox{\linewidth}{!}{
\begin{tabular}{p{2.3cm}p{5.6cm}cc}
\toprule
\textbf{Theme} & \textbf{Sub-themes} & \textbf{Seg.} & \textbf{\%} \\
\midrule
Heuristics \& Reasoning Patterns
  & Balancing act; Cognitive ease \& fluency;
    Selecting the lesser evil; Privacy vs.\ utility;
    Affective/intuitive heuristics
  & 157 & 30.9 \\
\addlinespace
User Perception of Privacy
  & Preference for targeted entity masking;
    Perceived privacy vulnerabilities;
    Unnecessary privatization;
    Failure in privatization
  & 115 & 22.6 \\
\addlinespace
Meaning \& Narrative Preservation
  & Retention of specific details;
    Faithfulness as constraint;
    Meaning over grammar
  & 114 & 22.4 \\
\addlinespace
Difficulty of Privatized Texts
  & Severe comprehension loss;
    Severe semantic loss;
    Linguistic noise
  & 91  & 17.9 \\
\addlinespace
Context
  & Context-dependent privacy tolerance;
    Appropriateness \& toxicity avoidance;
    Preservation of sentiment \& tone;
    Pragmatic failure
  & 31  & 6.1  \\
\bottomrule
\end{tabular}
}
\caption{Themes and sub-themes identified from 508 segments in our reflexive TA of 211 feedback
responses. Sub-themes are in order of associated segments (\textit{Seg.)}.}
\label{tab:themes_summary}
\end{table}

\paragraph{Thematic analysis.}
Following the methodology of \citet{braun2006using}, we conduct reflexive thematic analysis (TA) \cite{braun2019reflecting} of the freeform feedback responses from the survey. These responses offer a window into how users reason about privatized text, providing the opportunity for insights beyond the pure preference responses.

We followed the standard six-point process of reflexive TA, namely (1) familiarization via reading, (2) open coding of meaningful segments, (3) initial theme generation, (4) theme review, (5) theme labeling, and (6) finalization. Two researchers worked independently on steps 1-3, and collaborated for aggregation and theme finalization.

The full codebook resulting from our reflexive TA is found in Table \ref{tab:themes_full} of the Appendix, and an abridged version is presented in Table \ref{tab:themes_summary}. We identified 508 relevant segments from the 211 feedback texts, and categorized these into five overarching themes. To mitigate researcher bias, we compared our derived themes against the outputs of a \textsc{BERTopic} \cite{grootendorst2022bertopicneuraltopicmodeling} pipeline. This is detailed in Appendix \ref{sec:bertopic}, which shows a consistent clustering of the extracted TA themes.

The dominant theme, \textit{Heuristics and Reasoning Patterns}, describes the strategies used when weighing perturbed options, from a \say{balancing act} across meaning, readability, and privacy to selecting the \say{lesser evil} from sub-optimal options. \textit{User Perception of Privacy} describes perceptions at odds with DP text privatization, where privacy is often characterized as targeted entity masking. \textit{Meaning and Narrative Preservation} captures a clear prioritization of coherent meaning over privacy and a general rejection of violations such as inverted facts. \textit{Difficulty of Privatized Texts} describes adversity to noisy outputs, where lack of comprehensibility deters positive perceptions. \textit{Context} captures the intuitive notion that acceptable privatization depends on the sensitivity of the source material. Overall, these themes suggest that current DP text privatization is far from human acceptable, motivating the need for technical solutions to incorporate human feedback to bolster acceptability. 

\section{Aligning Models for DP Text Privatization: A Case Study}
\label{sec:case}
Motivated by integrating human privatization preferences in DP text privatization methods for more acceptable private outputs, we conduct a case study in language model alignment for DP text privatization. As our base case, we take the \textsc{DP-BART} \cite{igamberdiev-habernal-2023-dp} architecture. 

\paragraph{Motivation.}
Introduced in Section \ref{sec:construct}, \textsc{DP-BART} is realized via the addition of calibrated Gaussian noise to the latent representation between the encoder and decoder of a \textsc{BART} model \cite{lewis-etal-2020-bart}.
However, intuitively, decoding noisy latent representations will naturally produce sub-optimal outputs, since the decoder was pre-trained on clean representations. We therefore hypothesize that re-aligning the decoder of \textsc{DP-BART} will help to steer its generation process towards greater acceptability, in that alignment training rewards acceptable outputs over less acceptable alternatives.

\begin{figure}[t]
    \centering
    \includegraphics[trim=0.5cm 0.5cm 0.5cm 0.5cm, clip, width=\linewidth]{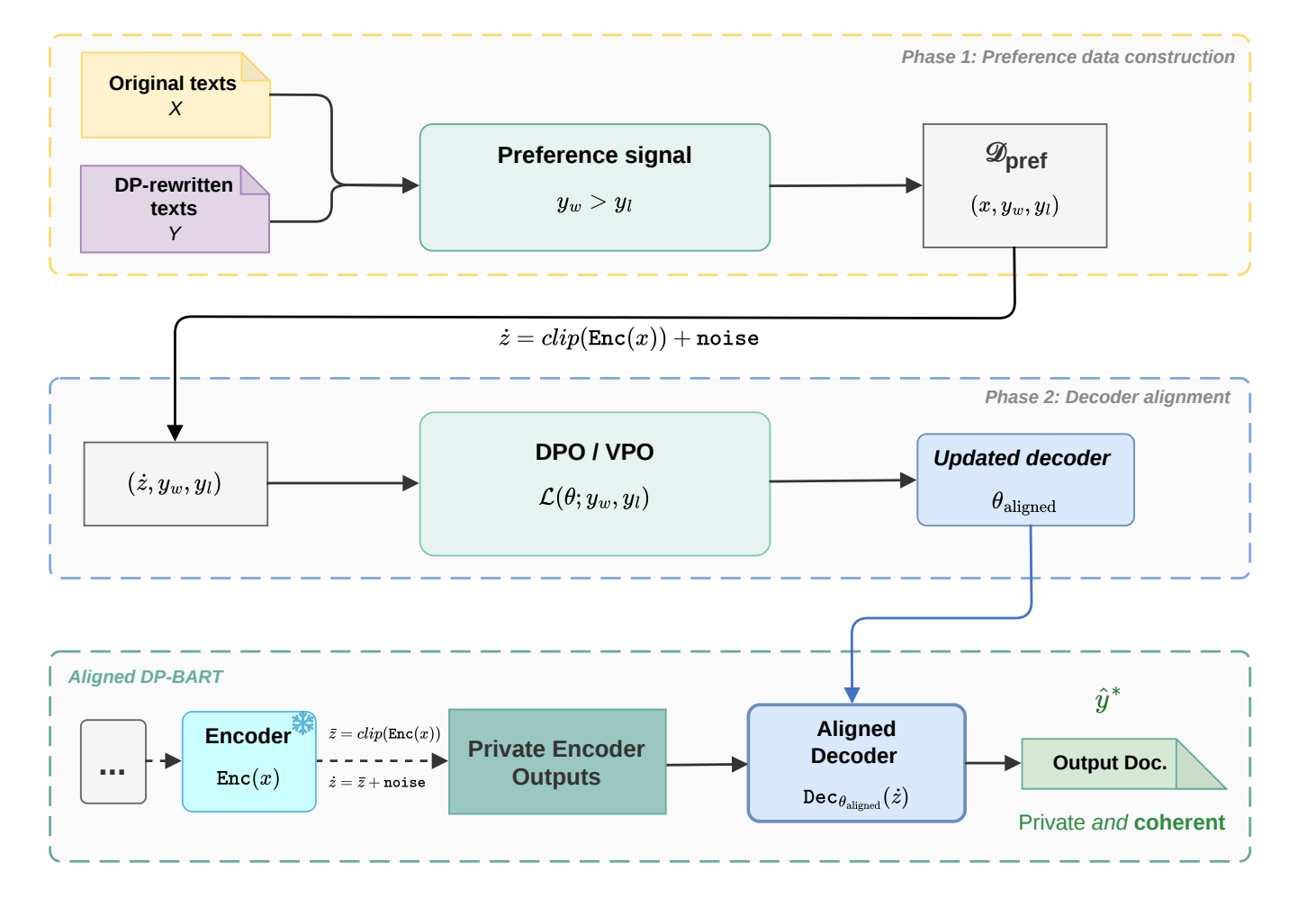}
    \caption{Aligning DP-BART with preference data.}
    \label{fig:pipeline_overview}
\end{figure}

\subsection{Alignment Training Setup}
We test three alignment pipelines: one based solely on human preferences, and two with augmented data using metric- and LLM-based expansion.

\paragraph{Human-anchored alignment.}
Configuration 1 constructs binary preference pairs from the survey responses (1080 question variants), presenting a theoretical limit of 2160 preference pairs (three options = two pairs). In \textbf{Configuration 1.1}, only winning preference with a strict majority (at least three votes) were included, yielding 1760 training pairs. This configuration uses pure DPO training \cite{rafailov2023direct} of the \textsc{BART} decoder. 

We also design \textbf{Configuration 1.2} to recover more nuanced preferences using the actual vote counts from the survey. For a variant with vote tallies of C = 3, B = 1, A = 0, this yields three pairs: C $\succ$ B, C $\succ$ A, and B $\succ$ A. The third pair would be absent from Configuration 1.1, but 1.2 retains these. As a result of this, an additional 827 pairs are recovered, bringing the total to 2,613 out of a possible 3,240. Each pair is annotated with a soft preference probability $\hat{p}$ derived from its vote counts using a Bayesian MMSE estimator\footnote{For vote counts $n_w$ and $n_l$, $\hat{p} = \frac{n_w + 1}{n_w + n_l +2}$.}. For example, a 4-0 vote yields $\hat{p} = \frac{4+1}{4+2} \approx 0.83$, while a 3-1 vote yields $\hat{p} = \frac{3+1}{4+2} \approx 0.67$. Training follows the VPO objective \cite{cho2024vpoleveragingnumbervotes}, which weights the loss for each pair by $\hat{p}$.

\paragraph{Preference data augmentation.}
As pure human-anchored alignment is limited by the collected survey data, we explore four methods to expand the preference data while also maintaining its signal.

In \textbf{Configuration 2.1}, we leverage the full set of 50 privatized texts per reference from the candidate scoring process of Section \ref{sec:construct}. We construct a preference pair for each of the candidate texts for which the majority-selected survey text holds a higher acceptability score. This gives a theoretical bound of 1080 $\times$ 49 = 52,920 pairs. However, since the top (100\textsuperscript{th} percentile) option was not always the majority winner, the actual number of preference pairs for this configuration was 34,022.

In \textbf{Configuration 2.2}, we only keep preference pairs with a distinct gap in acceptability score, as opposed to \textit{all} lesser scoring pairs in 2.1. We test two sub-configurations: all preferences pairs kept that are one standard deviation below the majority-chosen survey response text (2.2.1, \textit{relaxed}), and all those two standard deviations below \underline{and} where the non-preferred texts are at least one standard deviation below the mean (2.2.2, \textit{strict}).

In \textbf{Configuration 2.3}, we leverage an LLM to perform synthetic expansion of the base 120 texts. This is motivated by our post-survey analysis, which revealed that LLMs mirror human acceptability preferences (Table \ref{tab:model_tertile_toprank}). We selected \textsc{gpt-4o-mini} \cite{openai2024gpt4ocard}, which achieved the highest pairwise agreement with the human majority vote ($\alpha$ = 0.348, 62.0\% overall agreement) and the highest top-rank agreement (61.0\%). We replicated the survey process for three dataset sizes, i.e., across three methods $\times$ three privacy budgets: 120 (i.e., the original), 720 (12,960 preference pairs), and 1800 (32,400 preference pairs). The preference pair construction follows Configuration 1.1, where two preference pairs are constructed from each base LLM response (from three options).

Finally, in \textbf{Configuration 3.1}, an LLM-based \say{gold} privatized output (see prompt in Figure \ref{fig:anchor_prompt} of the Appendix) is anchored as the preferred pair against \textit{all} 50 \say{noisy} DP outputs. The prompt is grounded in the themes surfaced during the thematic content analysis, enabling a direct integration of human preference for acceptable privatized outputs. This yields preference pairs of the form ${(c_{LLM} \succ c_1), (c_{LLM} \succ c_2), \cdots, (c_{LLM} \succ c_{50})}$. We use only the 120 base texts, yielding 120 $\times$ 9 $\times$ 50 = 54,000 training pairs in total.

Further details on the alignment training setup and parameters can be found in Appendix \ref{sec:alignment}

\begin{figure*}[!t]
    \centering
    \includegraphics[width=\linewidth]{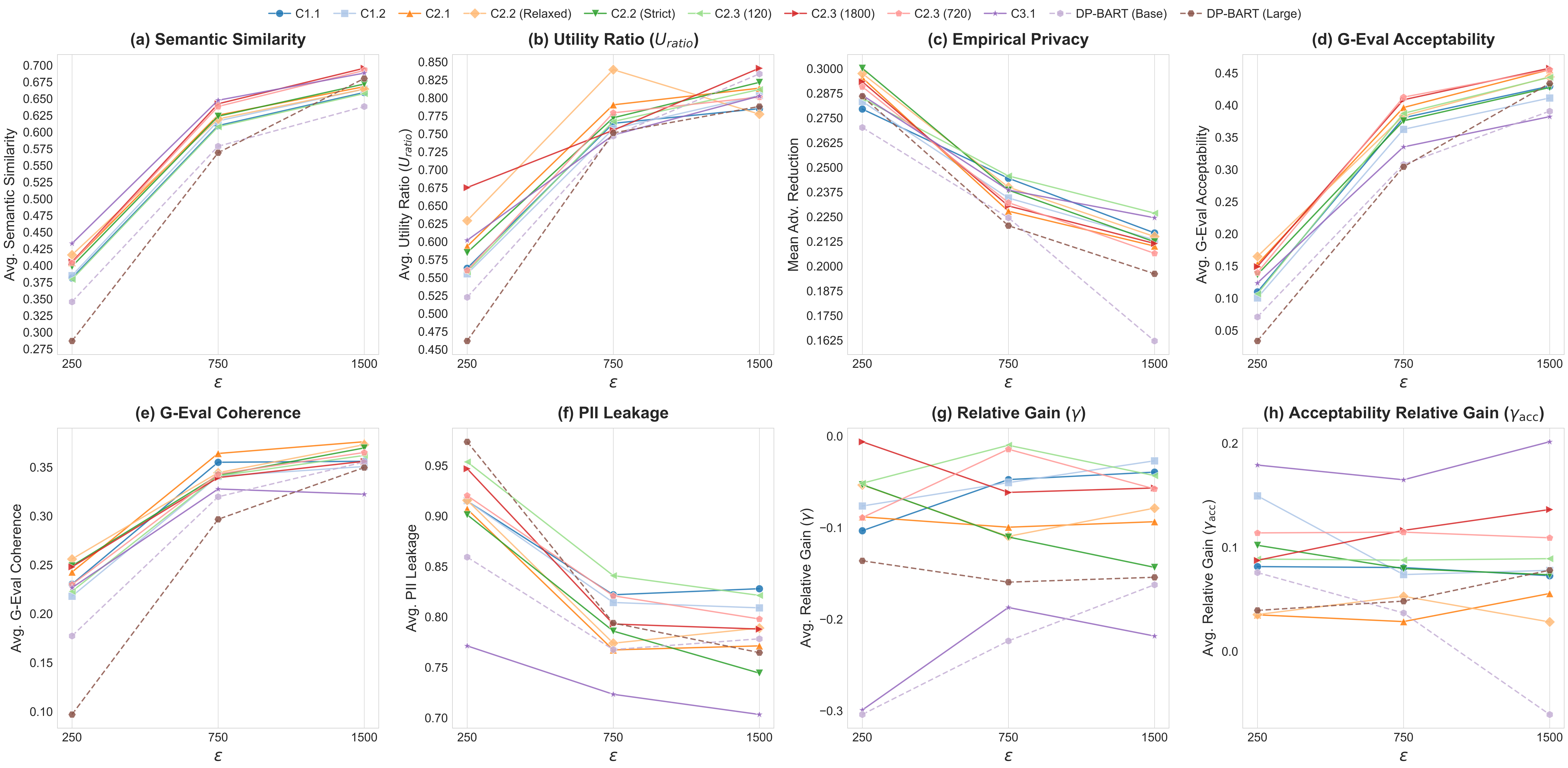}
    \caption{Evaluation results for selected metrics, averaged across all datasets. Higher scores are better in all panels.}
    \label{fig:results}
\end{figure*}

\subsection{Evaluation Procedure}
\label{sec:eval}
We evaluate all nine aligned \textsc{DP-BART} models (Configurations 1-3) against two \textsc{DP-BART} baselines, using the \textsc{base} and \textsc{large} \textsc{BART} models. We select three evaluation privacy budgets $\varepsilon \in \{250, 750, 1500\}$, i.e., each dataset (below) is privatized under three privacy regimes, yielding 27 private counterparts per dataset from our aligned models, in addition to the six baselines.

We use three datasets for evaluation, selected as a representative sample from our ten original datasets: Reddit Mental Health Posts, Trustpilot Reviews, and Yelp Reviews. Details on dataset characteristics and pre-processing steps are found in Appendix \ref{sec:appdatasets}. 
All three datasets contain a sensitive attribute (author or gender label), allowing for an adversarial inference task. Trustpilot and Yelp are marked with sentiment labels, enabling downstream utility preservation measurement.

In addition to downstream utility and empirical privacy, we measure the average semantic (cosine) similarity between input and private output text pairs. We also use G-Eval \cite{liu-etal-2023-g} to measure three quantities: text coherence, acceptability, and PII leakage.
Finally, we calculate four composite scores: utility ratio (average of downstream task performance, coherence, and similarity), privacy ratio (defense against adversarial inference and PII leakage), privacy-utility trade-off, and privacy-acceptability trade-off. The trade-offs are reported as \textit{relative gains}. Metric implementations are explained in greater detail in Appendix \ref{sec:appmetrics}.

\paragraph{Results.}
Our results are summarized in Figure \ref{fig:results}; the complete results are found in Appendix \ref{sec:complete}.

As can be seen, across the board, alignment training leads to improvements in privacy, utility, \textit{and} acceptability. DP outputs from our aligned models represent texts that are semantically more faithful to the original, comparably or better in terms of empirical privacy protection, and most importantly, higher in coherence and acceptability. This is ultimately reflected in the calculated trade-off scores, where our aligned models reach closest to parity in privacy-utility trade-offs ($\gamma$) and score the highest in acceptability-privacy trade-offs ($\gamma_{\text{acc}}$).

Building on these findings, we run statistical \textit{t}-tests to validate the significance of utility and acceptability gains. At the macro level (pooling all nine aligned configurations vs. pooled baselines), the aligned group significantly outperforms on semantic similarity (Cohen’s \textit{d} = 0.647) and metric acceptability (\textit{d} = 0.754), both with medium-to-large effects. Utility is statistically significant but negligible in magnitude (\textit{d} = 0.027). This effect, however, differs based on privacy regime. At $\varepsilon$ = 250, semantic similarity reaches \textit{d} = 1.114 and metric acceptability \textit{d} = 0.763. At $\varepsilon$  = 1500, these shrink to \textit{d} = 0.069 and \textit{d}= 0.380, respectively.

A micro comparison by configuration family confirms that all three setups significantly outperform the baselines on semantic similarity and metric acceptability. Full per-configuration statistical results are in Table \ref{tab:app_param_ttests} of Appendix \ref{sec:complete}.

\subsection{The Value of Alignment}
We compare our aligned models to non-aligned counterparts, i.e., via using sequence-to-sequence models for post-processing DP texts \cite{10.1145/3664476.3669926}. In this, we measure the value of alignment, and whether this outperforms simpler, non-human-informed post-processing.

We leverage two models for baseline post-processing: \textsc{flan-t5-base} and \textsc{flan-t5-large} \cite{3722577.3722647}, with the  prompt \textit{Correct and improve the grammar, acceptability, and coherence of the following text: \{DP TEXT\}}. We limited the input length to 256 tokens, the generated output to 128 tokens, and used greedy decoding with \texttt{num\_beams}=2. This process yielded 36 additional dataset counterparts (2 DP methods [$base$ and $large$] $\times$ 3 datasets $\times$ 3 $\varepsilon$ $\times$ 2 post-processing models). These were evaluated on the same acceptability metrics\footnote{With the caveat of G-Eval, which was run with \textsc{gpt-5.6-luna}, as the previously used \textsc{gpt-5.4-nano} was deprecated.} as presented in Section \ref{sec:eval}.

\begin{table*}[htbp]
\centering
\small
\resizebox{0.97\textwidth}{!}{
\begin{tabular}{ll|ccccc|ccccc|ccccc}
\toprule
& & \multicolumn{5}{c|}{$\epsilon = 250$} & \multicolumn{5}{c|}{$\epsilon = 750$} & \multicolumn{5}{c}{$\epsilon = 1500$} \\
\cmidrule(lr){3-7} \cmidrule(lr){8-12} \cmidrule(lr){13-17}
\textbf{Dataset} & \textbf{Method} & \textbf{SS} & \textbf{COH} & \textbf{PII} & \textbf{GE$_{\text{acc}}$} & \textbf{MA} & \textbf{SS} & \textbf{COH} & \textbf{PII} & \textbf{GE$_{\text{acc}}$} & \textbf{MA} & \textbf{SS} & \textbf{COH} & \textbf{PII} & \textbf{GE$_{\text{acc}}$} & \textbf{MA} \\
\midrule
\multirow{5}{*}{\textbf{Reddit Mental Health}} & \textbf{Best Aligned (Ours)} & \textbf{0.455} & 0.258 & 0.963 & \textbf{0.198} & -0.745 & \textbf{0.668} & 0.342 & 0.629 & \textbf{0.450} & -0.505 & \textbf{0.708} & 0.324 & 0.626 & \textbf{0.500} & -0.456 \\
\cmidrule(lr){2-17}
 & \textsc{DP-BART-base} + \textsc{flan-t5-base} & 0.344 & 0.272 & 0.950 & 0.027 & -0.501 & 0.530 & \textbf{0.538} & 0.613 & 0.210 & -0.469 & 0.575 & \textbf{0.696} & 0.556 & 0.180 & -0.437 \\
 & \textsc{DP-BART-large} + \textsc{flan-t5-base} & 0.289 & \textbf{0.276} & \textbf{1.000} & 0.017 & -0.646 & 0.523 & 0.412 & 0.674 & 0.150 & -0.437 & 0.613 & 0.656 & \textbf{0.650} & 0.273 & -0.367 \\
 & \textsc{DP-BART-base} + \textsc{flan-t5-large} & 0.352 & 0.238 & 0.900 & 0.017 & \textbf{-0.415} & 0.576 & 0.444 & 0.543 & 0.170 & -0.333 & 0.623 & 0.564 & 0.550 & 0.212 & -0.291 \\
 & \textsc{DP-BART-large} + \textsc{flan-t5-large} & 0.290 & 0.234 & \textbf{1.000} & 0.017 & -0.600 & 0.560 & 0.364 & \textbf{0.714} & 0.225 & \textbf{-0.330} & 0.655 & 0.532 & 0.550 & 0.320 & \textbf{-0.269} \\
\midrule
\multirow{5}{*}{\textbf{Trustpilot}} & \textbf{Best Aligned (Ours)} & \textbf{0.420} & 0.273 & 0.956 & \textbf{0.177} & -0.946 & \textbf{0.661} & 0.391 & \textbf{0.982} & \textbf{0.496} & -0.564 & \textbf{0.721} & 0.432 & 0.972 & \textbf{0.554} & -0.492 \\
\cmidrule(lr){2-17}
 & \textsc{DP-BART-base} + \textsc{flan-t5-base} & 0.338 & 0.260 & 0.822 & 0.020 & -0.724 & 0.537 & 0.602 & 0.800 & 0.455 & -0.554 & 0.590 & \textbf{0.708} & 0.975 & 0.417 & -0.446 \\
 & \textsc{DP-BART-large} + \textsc{flan-t5-base} & 0.290 & 0.216 & \textbf{1.000} & 0.022 & -0.868 & 0.537 & \textbf{0.686} & 0.900 & 0.395 & -0.517 & 0.645 & 0.618 & 0.900 & 0.508 & -0.343 \\
 & \textsc{DP-BART-base} + \textsc{flan-t5-large} & 0.342 & \textbf{0.342} & 0.767 & 0.022 & \textbf{-0.632} & 0.566 & 0.502 & 0.800 & 0.448 & \textbf{-0.427} & 0.626 & 0.576 & 0.880 & 0.465 & -0.318 \\
 & \textsc{DP-BART-large} + \textsc{flan-t5-large} & 0.290 & 0.210 & \textbf{1.000} & 0.033 & -0.818 & 0.559 & 0.576 & 0.887 & 0.398 & -0.430 & 0.671 & 0.606 & \textbf{1.000} & 0.532 & \textbf{-0.264} \\
\midrule
\multirow{5}{*}{\textbf{Yelp}} & \textbf{Best Aligned (Ours)} & \textbf{0.424} & 0.268 & \textbf{0.992} & \textbf{0.121} & -0.838 & \textbf{0.622} & 0.391 & 0.972 & \textbf{0.302} & \textbf{-0.642} & \textbf{0.671} & 0.416 & 0.981 & 0.350 & \textbf{-0.589} \\
\cmidrule(lr){2-17}
 & \textsc{DP-BART-base} + \textsc{flan-t5-base} & 0.323 & 0.264 & 0.983 & 0.057 & -0.604 & 0.474 & \textbf{0.594} & \textbf{1.000} & 0.135 & -0.771 & 0.413 & 0.458 & \textbf{1.000} & 0.092 & -1.294 \\
 & \textsc{DP-BART-large} + \textsc{flan-t5-base} & 0.279 & 0.190 & 0.900 & 0.028 & -0.589 & 0.434 & 0.452 & 0.967 & 0.090 & -1.093 & 0.604 & \textbf{0.570} & 0.735 & 0.398 & -0.689 \\
 & \textsc{DP-BART-base} + \textsc{flan-t5-large} & 0.330 & \textbf{0.290} & 0.967 & 0.037 & \textbf{-0.507} & 0.519 & 0.530 & 0.875 & 0.175 & -0.657 & 0.433 & 0.372 & 0.950 & 0.083 & -1.249 \\
 & \textsc{DP-BART-large} + \textsc{flan-t5-large} & 0.279 & 0.202 & 0.900 & 0.025 & -0.535 & 0.449 & 0.400 & \textbf{1.000} & 0.070 & -1.053 & 0.635 & 0.464 & 0.555 & \textbf{0.442} & -0.612 \\
\bottomrule
\end{tabular}
}
\caption{Comparison of metrics from our best aligned score across configurations~1.1--3.1 against simple, post-hoc rewriting baselines, across all datasets and privacy budgets ($\epsilon$). Higher is better for all reported metrics.}
\label{tab:flan_postproc_comparison}
\end{table*}

The results are provided in Table \ref{tab:flan_postproc_comparison}, compared to the per-metric best values from our alignment configurations. As shown, our alignment efforts always produce outputs that are significantly more semantically similar (SS) to the originals, and nearly always more acceptable (according to G-Eval). PII defense is comparably strong across the board.

On the surface, we see that simple post-processing often produces higher G-Eval coherence (COH) scores, as well as more favorable metric acceptability. We hypothesize, however, that this is largely due to output length. On average, the simple post-processing produced significantly shorter texts than the original (39.11 words, or 61.3\% of the original length). Thus, we find that acceptability measurements of privatized texts are susceptible to length bias (especially non-LLM based metrics), thus suggesting a point for future improvements.

\section{Discussion}
\paragraph{Taming noise.}
Our work is rooted in addressing the adverse effects of noisification in DP text privatization. This, although necessary for achieving DP, yields unpredictable and often unacceptable privatized outputs, particularly at stricter $\varepsilon$. The challenge is only exacerbated by building text privatization mechanisms with components not well-equipped to handle noisy representations, as with \textsc{DP-BART}. Our proposed solution, however, is to leverage the variety introduced by noisification, which lends itself well to collecting preferences -- what is \say{acceptable} privatization versus what is not. Such a pursuit must be grounded in some quantifiable notion of acceptability, which we propose to be a composite of multiple \textit{facets} from the existing literature on natural language evaluation.

\paragraph{The merit in human-centered text privatization.}
The collection of privatization preferences is insightful (as shown with our TA), yet it does not fully address the usability challenge of DP text privatization. In order to harness the signal contained within these preferences, we show that existing language model alignment frameworks can be leveraged to steer privatization mechanisms back towards acceptability. Furthermore, we empirically demonstrate that doing so presents many advantages: higher coherence and acceptability, but also increased utility, privacy, and trade-offs. Such results, replicated across privacy budgets and evaluation datasets, make a compelling case for human-informed DP text privatization, and our framework enables this with greater acceptability and usability. 

\section{Conclusion}
We address the challenge of \textit{acceptability} in DP text privatization by aligning language model-based mechanisms with human text privatization preferences. We ground our study in a review of 103 literature sources, providing the basis for an operationalization of acceptability. Leveraging this, we construct a text privatization preference dataset and survey 294 humans from around the globe to build preference pairs for alignment training. In a series of alignment training configurations, we demonstrate that DP mechanisms built on top of aligned models outperform baselines across an array of metrics, privacy budgets, and evaluation datasets. Our work lays the foundation for one pathway towards more acceptable and usable DP text privatization, grounded in human privacy perceptions.

\paragraph{Outlook.}
We see two important paths for continued research. Firstly, we call for increased work to integrate human privacy perceptions into text privatization mechanisms, which includes improving methodologies to gather such perceptions accurately and holistically. Accordingly, we argue for a higher concentration on \textit{usable} DP privatization, which, as shown, can benefit from modeling human perceptions of acceptable privatization outcomes.

\section*{Limitations}
The first limitation of our work, particularly in relation to the conclusions we draw, is that the observed benefits and performance improvements do not come strictly for free. Alignment training adds substantial overhead over standard \textsc{DP-BART} inference (although it is one-time), and the human preference collection is relatively costly. Our survey budget was near $\pounds$900, and scaling to the dataset sizes that produced the strongest training signals in this study would multiply
that figure considerably. We show, however, that with a strong basis in human preferences, LLM proxies can be used more heavily to scale, at a negligible comparative cost. 

As with any human-based assessment (i.e., here, our survey), and particularly in the arena of \textit{privacy}, answers may be noisy due to task misunderstanding, survey fatigue, or other environmental factors. We especially acknowledge that the understanding of \textit{what makes a text private?} (let alone acceptable) is a difficult question that still pervades research. We mitigated this limitation by building in a redundancy factor (each text is reviewed by four survey participants), by incorporating clear instructions (and piloting the survey), and by validating the response distribution with LLM-as-a-Judge. Despite this, we cannot and do not rule out that some of the collected responses may be noisy, and we encourage the interpretation of our results (and the released preference datasets) as such.

We also draw attention to our chosen evaluation procedure, which is rooted in proxy measures of privacy, utility, and acceptability. We particularly acknowledge the limitations of \textsc{G-Eval}, which has been shown to be susceptible to positional bias \cite{3666122.3668142}, ordering bias \cite{wang-etal-2024-large-language-models-fair}, and scoring instability \cite{raina-etal-2024-llm}. Nevertheless, we mitigate both of these limitations by selecting a diverse array of evaluation metrics, both LLM- and non-LLM based, and additionally, no single evaluation focus (privacy, utility, acceptability) relies solely on LLM-based assessments. 

Finally, we acknowledge that our case study and alignment framework are tailored specifically to \textsc{DP-BART} and encoder-decoder-based DP text privatization. Adapting this framework to use preference data to align other, mechanistically different DP methods would require further work, and this may yield different empirical results. Here, we highlight the main theoretical contributions in the literature review, dataset, and survey we establish and construct, upon which our case study is built. We hope that this serves as a starting point for future validation and extension studies.

\section*{Ethical Considerations}
In the administration of our online survey to 294 human participants across the globe, we ensured a fair compensation rate of $\pounds$9 per hour, marked as a fair rate by Prolific. We also ensured the well-being of the participants by limiting the required responses (10 questions to answer with an estimated 15-minute total duration), as well as by screening out overtly sensitive or vulgar reference texts in our preference dataset construction process. We ensured informed consent clearly outlining the study purpose, time commitment, and data use (for research purposes), and we did not collect any personally identifiable information.

All utilized datasets, both for preference data and for evaluation, as well as all utilized software and models, are open-source and publicly available. Furthermore, our use of these resources is in line with their associated licenses.

We foresee, to the best of our knowledge, no negative or adversarial uses of our proposed work (metrics, dataset, alignment framework, and aligned models). As such, our full replication package is released under a permissive MIT license.



\bibliography{custom}

\appendix

\section{Supplemental Material}
\label{sec:appendix}

\subsection{Literature Review}
\label{sec:litreview}
\paragraph{Inclusion and exclusion criteria}
Below are included the three inclusion (I) and six exclusion (E) criteria that guided the selection process in our scoping literature review.

\begin{itemize}
    \small
    \itemsep 0em
    \item[I1.] The study proposes or evaluates a mechanism for (DP) text privatization, including applications in summarization, question answering, dialogue systems, etc.
    \item [I2.] The study conducts empirical evaluation using at least one quantitative metric or human evaluation.
    \item[I3.] Peer-reviewed, or preprint with more than 10 citations or released within the past 12 months.
\end{itemize}

\begin{itemize}
    \small
    \itemsep 0em
    \item[E1.] Focuses solely on non-textual modalities (tabular data, images, time-series)
    \item[E2.] No text generation involved; classification or embedding training only
    \item[E3.] Purely theoretical contribution without empirical utility assessment
    \item[E4.] Evaluation limited to downstream task accuracy without intrinsic text quality metrics
    \item[E5.] Paper not available in English
    \item[E6.] Method not designed or evaluated for English
\end{itemize}

In Table \ref{tab:metrics-scoping-review}, we provide the complete collections of metrics extracted from our scoping literature. These informed the representative sample discussed in the work and found in Table \ref{tab:chosen-metrics}.

\begin{table*}[ht]
\small
\centering
\resizebox{\textwidth}{!}{%
\begin{tabular}{p{0.25\textwidth}ccp{0.9\textwidth}}
\toprule
\textbf{Metric} & \textbf{Granularity} & \textbf{Type} & \textbf{Key Characteristic} \\
\midrule

\multicolumn{4}{l}{\textit{Semantic similarity}} \\
\midrule
BERTScore \cite{zhang2020bertscore} & Token-level & Ref-based & Contextual embedding matching (cosine) between candidate and reference tokens; robust to paraphrase. \\
BLEU \cite{papineni-etal-2002-bleu} & Token-level & Ref-based & Precision-oriented n-gram overlap; widely used baseline but poorly correlated with human judgment. \\
ROUGE \cite{lin-2004-rouge} & Token-level & Ref-based & Recall-oriented n-gram and longest common subsequence overlap family (ROUGE-N, ROUGE-L). \\
METEOR \cite{banerjee-lavie-2005-meteor} & Token-level & Ref-based & Word-alignment with stemming and synonym lookup (WordNet); improves correlation with human judgments over BLEU. \\
CIDEr \cite{7299087} & Token-level & Ref-based & TF-IDF weighted n-gram consensus; penalizes common words and emphasizes distinctive information. \\
ChrF \cite{popovic-2015-chrf} & Token-level & Ref-based & Character-level n-gram F-score; effective for morphologically rich languages and noisy text. \\
NIST \cite{1289189.1289273} & Token-level & Ref-based & Improves on BLEU by weighting rare (informative) n-grams more heavily. \\
GLEU \cite{napoles-etal-2015-ground} & Token-level & Ref-based & BLEU variant accounting for both source and reference; optimized for sentence-level grammatical error correction. \\
BEER \cite{stanojevic-simaan-2014-beer} & Token-level & Ref-based & Uses linguistically motivated features (unigrams, POS tags, reordering patterns) with a pairwise ranking model. \\
SARI \cite{xu-etal-2016-optimizing} & Token-level & Ref-based & Measures goodness of keep, add, and delete operations; designed for text simplification evaluation. \\
TER \cite{snover-etal-2006-study} & Token-level & Ref-based & Counts edits (insertions, deletions, shifts) needed to transform output into the reference. \\
CDER \cite{leusch-etal-2006-cder} & Token-level & Ref-based & Edit distance metric that efficiently handles block movements and reordering. \\
WER \cite{niessen-etal-2000-evaluation} & Token-level & Ref-based & Strict Levenshtein distance at the word level; often too harsh for generative tasks. \\
Greedy Matching \cite{rus-lintean-2012-comparison} & Token-level & Ref-based & Greedily aligns each generated word to a reference word based on embedding cosine similarity. \\
BERT-iBLEU \cite{sun-zhou-2012-joint} & Token-level & Ref-based & Hybrid metric combining BERT contextual embeddings with BLEU precision. \\
FKBLEU \cite{xu-etal-2016-optimizing} & Token-level & Ref-based & Combines Flesch-Kincaid readability features with iBLEU; designed for text simplification evaluation. \\
GTM \cite{turian-etal-2003-evaluation} & Token-level & Ref-based & F-score based on precision and recall with greater weight for contiguous matching spans. \\
SAMSA \cite{sulem-etal-2018-semantic} & Document-level & Ref-free & Leverages semantic parsing to evaluate structural simplification quality; reference-less. \\
BLEURT \cite{sellam-etal-2020-bleurt} & Document-level & Ref-based & Learned BERT-based regression metric fine-tuned on human judgments; strong correlation with human quality scores. \\
COMET \cite{rei-etal-2020-comet} & Document-level & Ref-based & Neural evaluation framework trained on human judgments; cross-encoder analyzing source, hypothesis, and reference. \\
COMETKiwi \cite{rei-etal-2022-cometkiwi} & Document-level & Ref-based & Quality estimation variant of COMET for sentence-level assessment without references at inference. \\
BARTScore \cite{3540261.3542349} & Document-level & Ref-based & Formulates evaluation as a generation probability task; can assess informativeness, fluency, or factuality. \\
PRISM \cite{thompson-post-2020-automatic} & Document-level & Ref-based & Frames evaluation as scoring output with a multilingual sequence-to-sequence paraphraser. \\
RUSE \cite{shimanaka-etal-2018-ruse} & Document-level & Ref-based & Regressor using multiple sentence embedding types to predict translation quality. \\
YiSi \cite{lo-2019-yisi} & Document-level & Ref-based & Measures semantic similarity using weighted lexical embeddings across different resource levels. \\
PARENT \cite{dhingra-etal-2019-handling} & Document-level & Ref-based & Aligns n-grams from reference and generated text to semi-structured data before computing precision and recall. \\
Sentence Mover's Similarity \cite{clark-etal-2019-sentence} & Document-level & Ref-based & Extends Word Mover's Distance to sentence-level embeddings to capture logical flow. \\
Embedding Average \cite{wieting2016universalparaphrasticsentenceembeddings} & Document-level & Ref-based & Cosine similarity of averaged word vectors (GloVe/Word2Vec); fast but crude baseline. \\
Embedding Cosine Sim. \cite{reimers-gurevych-2019-sentence} & Document-level & Ref-free & Cosine similarity between dense sentence embeddings from pre-trained encoders. \\
SPICE \cite{anderson2016spice} & Document-level & Ref-based & Scene-graph based semantic propositional evaluation; focuses on semantic content over surface overlap. \\
MoverScore \cite{zhao-etal-2019-moverscore} & Corpus-level & Ref-based & Uses contextual embeddings with Earth Mover's Distance to measure semantic transport cost. \\

\midrule
\multicolumn{4}{l}{\textit{Fluency}} \\
\midrule
Perplexity (PPL) \cite{jelinek1977perplexity} & Token-level & Ref-free & Language model log-likelihood measuring how predictable text is; proxy for fluency. \\
ENTF \cite{yu-etal-2015-casict} & Token-level & Ref-based & Entropy-based method introducing unigram F-score for fluency evaluation. \\
DPMF \cite{yu-etal-2015-casict} & Token-level & Ref-based & Combined metric incorporating lexical, syntactic, and semantic features using dependency parsing. \\
CoLA Classifier \cite{warstadt2019neural} & Document-level & Ref-free & Binary classification of grammatical acceptability (well-formed vs. broken). \\
SLOR \cite{kann-etal-2018-sentence} & Document-level & Ref-free & Syntactic log-odds ratio; normalizes perplexity by unigram probability to isolate syntactic structure. \\
Scribendi Score \cite{islam-magnani-2021-end} & Document-level & Ref-free & Reference-free GEC metric based on GPT-2 perplexity scores. \\
SESCORE \cite{xu-etal-2022-errors} & Document-level & Ref-free & Learned metric pipeline producing nuanced quality scores without human-annotated reference pairs. \\
G-Eval \cite{liu-etal-2023-g} & Document-level & Ref-free & LLM-as-a-judge framework using chain-of-thought prompting to score coherence, fluency, etc. \\
USR \cite{mehri-eskenazi-2020-usr} & Document-level & Ref-free & Unsupervised reference-free metric measuring understandability, naturalness, context, and knowledge use in dialog. \\
MAUVE \cite{3540261.3540629} & Corpus-level & Ref-free & Distributional comparison between generated and human text using divergence frontiers; measures human-likeness. \\
Self-BLEU \cite{10.1145/3209978.3210080} & Corpus-level & Ref-free & Measures intra-sample repetition by comparing generated sentences against each other; detects mode collapse. \\
Distinct-N \cite{li-etal-2016-diversity} & Corpus-level & Ref-free & Fraction of unique n-grams; measures lexical diversity and detects repetitive loops. \\
Zipf Coefficient \cite{Holtzman2020The} & Corpus-level & Ref-free & Checks whether word frequency distribution follows Zipf's law as a proxy for distributional realism. \\
HUSE \cite{hashimoto-etal-2019-unifying} & Corpus-level & Ref-free & Combines automatic statistical metrics with human judgments via an optimal discriminator framework. \\
Coh-Metrix \cite{graesser2004coh} & Corpus-level & Ref-free & Suite of 200+ cohesion indices including connectives, referential overlap, and latent semantic analysis. \\
Lexical Deviation / Word Pair Deviation \cite{liu-soh-2022-towards} & Corpus-level & Ref-free & Measures how much vocabulary usage deviates from a standard norm or corpus. \\
DiscoScore \cite{zhao-etal-2023-discoscore} & Corpus-level & Ref-based & Models discourse coherence and focus between sentences using source-aware structure. \\

\midrule
\multicolumn{4}{l}{\textit{Readability}} \\
\midrule
Lexical Diversity (MTLD) \cite{mccarthy2010mtld} & Corpus-level & Ref-free & Measures vocabulary richness via type-token analysis; less sensitive to text length than basic TTR. \\
Coleman-Liau Index (CLI) \cite{coleman1975computer} & Document-level & Ref-free & Proxies complexity by character count; robust for noisy DP text where syllable counting breaks down. \\
Dale-Chall \cite{dale1948formula} & Document-level & Ref-free & Gauges difficulty by word familiarity (3,000 common-word list) and average sentence length. \\
Flesch-Kincaid Grade Level \cite{kincaid1975derivation} & Corpus-level & Ref-free & Uses sentence length and syllable count to estimate U.S. school grade level of text. \\
Automated Readability Index \cite{smith1967automated} & Corpus-level & Ref-free & Outputs grade level based on characters per word and words per sentence. \\
Gunning Fog Index \cite{gunning1952technique} & Corpus-level & Ref-free & Weights complex words (3+ syllables) heavily in its readability estimate. \\
SMOG \cite{mc1969smog} & Corpus-level & Ref-free & Estimates years of education needed based on polysyllabic word count in a 30-sentence sample. \\
Linsear Write \cite{klare1974assessing} & Corpus-level & Ref-free & Balances simple vs. complex words; originally designed for technical documentation. \\

\midrule
\multicolumn{4}{l}{\textit{Other}} \\
\midrule
FrugalScore \cite{kamal-eddine-etal-2022-frugalscore} & Document-level & Ref-free & Lighter alternative to BERTScore and MoverScore using distilled contextual embeddings. \\ 

\bottomrule
\end{tabular}
}
\caption{Overview of evaluation metrics identified in the scoping review.}
\label{tab:metrics-scoping-review}
\end{table*}

\subsection{Reference Datasets}
\label{sec:datasets}
Below, we outline the 10 reference datasets used for the preference dataset construction, summarizing their content and original curators.

\begin{itemize}[leftmargin=1em]
  \itemsep -0.3em
  \item \textbf{Blog Authorship Corpus}: Blog posts from \texttt{blogger.com} \cite{schler2006effects}.
  \item \textbf{Enron Emails}: Sent-folder emails from the Enron email dataset \cite{klimt2004enron}.
  \item \textbf{Medical Questions}: User-written questions from a large medical question-answering dataset\footnote{\scriptsize\href{https://hf.co/datasets/Malikeh1375/medical-question-answering-datasets}{Malikeh1375/medical-question-answering-datasets}}. Only the questions are used.
  \item \textbf{Mental Health Blog}: Posts from a mental health community \cite{boinepelli-etal-2022-leveraging}.
  \item \textbf{Reddit Confessions}: Personal confessions from a corpus of four confession subreddits\footnote{\scriptsize\href{https://hf.co/datasets/SocialGrep/one-million-reddit-confessions}{SocialGrep/one-million-reddit-confessions}}.
  \item \textbf{Reddit Legal Advice}: Requests for legal help on \texttt{/r/legaladvice} \cite{li-etal-2022-parameter}.
  \item \textbf{Reddit Mental Health Posts}: Posts from five mental health-related subreddits\footnote{\scriptsize\href{https://hf.co/datasets/solomonk/reddit_mental_health_posts}{solomonk/reddit\_mental\_health\_posts}}.
  \item \textbf{Trustpilot Reviews}: Reviews from the Trustpilot platform \cite{10.1145/2736277.2741141}.
  \item \textbf{Twitter}: A large corpus of tweets \cite{enryu_2023_15086029}.
  \item \textbf{Yelp Reviews}: A subset of user reviews from the Yelp platform \cite{utpala-etal-2023-locally}.
\end{itemize}

\subsection{Acceptability Metrics}
\label{sec:metrics}
The following outlines the calculation of our chosen acceptability metrics, as well as details how they are normalized and averaged to form our three facet scores (Equations \ref{eq:util}-\ref{eq:read}).

\paragraph{Semantic Fidelity.}
BERTScore \cite{zhang2020bertscore} is calculated using the Hugging Face \textsc{evaluate} default implementation\footnote{\url{https://github.com/huggingface/evaluate}}. BLEURT \cite{sellam-etal-2020-bleurt} is calculated using the official Google implementation\footnote{\url{https://github.com/google-research/BLEURT}}. Both scores are normalized via internal batch normalization (i.e., relative to the 50 candidate texts), since $\text{BERTScore}(\text{ref}, \text{ref}) = 1.0$ and $\text{BLEURT}(\text{ref}, \text{ref})$ is near-constant. Outliers are clipped at $\pm 3\sigma$ before averaging.
\begin{displaymath}
\centering
Z_{\text{BERT}}   = \frac{x_{i} - \mu_{\text{batch}}}{\sigma_{\text{batch}}},
Z_{\text{BLEURT}} = \frac{x_{i} - \mu_{\text{batch}}}{\sigma_{\text{batch}}}
\end{displaymath}

\begin{equation}
\label{eq:util}
\small
\text{Score}_{\text{Fidelity}} = \frac{\text{Clip}_{3\sigma}(Z_{\text{BERT}}) + \text{Clip}_{3\sigma}(Z_{\text{BLEURT}})}{2}
\end{equation}

\paragraph{Fluency.}

PPL and CoLA are normalized externally so that $0.0$ corresponds to the fluency of the original text. The raw CoLA score is calculated using \textsc{cointegrated/roberta-large-cola-krishna2020} \cite{krishna-etal-2020-reformulating}. Perplexity is calulcated using \textsc{GPT-2} \cite{radford2019language}, and the score is inverted before normalization ($x'_{\text{PPL}} = -x_{\text{PPL}}$), since lower perplexity indicates better fluency) and is not subject to additional rescaling, as it is the only metric in this facet outside the $[0,1]$ range.
\begin{displaymath}
\centering
Z_{\text{PPL}}  = \frac{x'_{\text{PPL}} - \mu_{\text{baseline, PPL}}}{\sigma_{\text{baseline, PPL}}}
\end{displaymath}

\begin{displaymath}
\centering
Z_{\text{CoLA}} = \frac{x_{\text{CoLA}} - \mu_{\text{baseline, CoLA}}}{\sigma_{\text{baseline, CoLA}}}
\end{displaymath}

\begin{equation}
\label{eq:flu}
\text{Score}_{\text{Fluency}} = \frac{Z_{\text{PPL}} + Z_{\text{CoLA}}}{2}
\end{equation}

\paragraph{Readability.}

DCRS and CLI areboth calculated using \textsc{textstat}\footnote{\url{https://pypi.org/project/textstat/}}. They are then z-scored against the reference baseline distribution, and the absolute deviation is taken to penalize stylistic drift in either direction and account for both text over-simplification and over-complexification relative to the original.
\begin{displaymath}
\centering
Z_{\text{DCRS}} = -\left|\frac{x_{\text{DCRS}} - \mu_{\text{baseline, DCRS}}}{\sigma_{\text{baseline, DCRS}}}\right|
\end{displaymath}

\begin{displaymath}
\centering
Z_{\text{CLI}}  = -\left|\frac{x_{\text{CLI}} - \mu_{\text{baseline, CLI}}}{\sigma_{\text{baseline, CLI}}}\right|
\end{displaymath}

\begin{equation}
\label{eq:read}
\text{Score}_{\text{Readability}} = \frac{Z_{\text{DCRS}} + Z_{\text{CLI}}}{2}
\end{equation}

\paragraph{Aggregated Score.}
For a final, aggregated \textit{Acceptability} score, we opt for a simple mean of the three component metrics, which have been z-scored along their reference distributions. The raw acceptability score is thus formalized as:
\begin{displaymath}
    \small
    A_{\text{raw}} = \frac{1}{3}(\text{Score}_{\text{Fidelity}} + \text{Score}_{\text{Fluency}} + \text{Score}_{\text{Readability}})
\end{displaymath}

From this, we reach the final acceptability score by rescaling to the range [0, 1] using the following:

\begin{displaymath}
    \text{Metric Acceptability} \ (A) = \frac{A_{\text{raw}} + 3}{6}
\end{displaymath}

\subsection{Privatization calibration.}
\label{sec:calibrate}
\begin{figure}[ht]
  \centering
  \includegraphics[width=\linewidth]{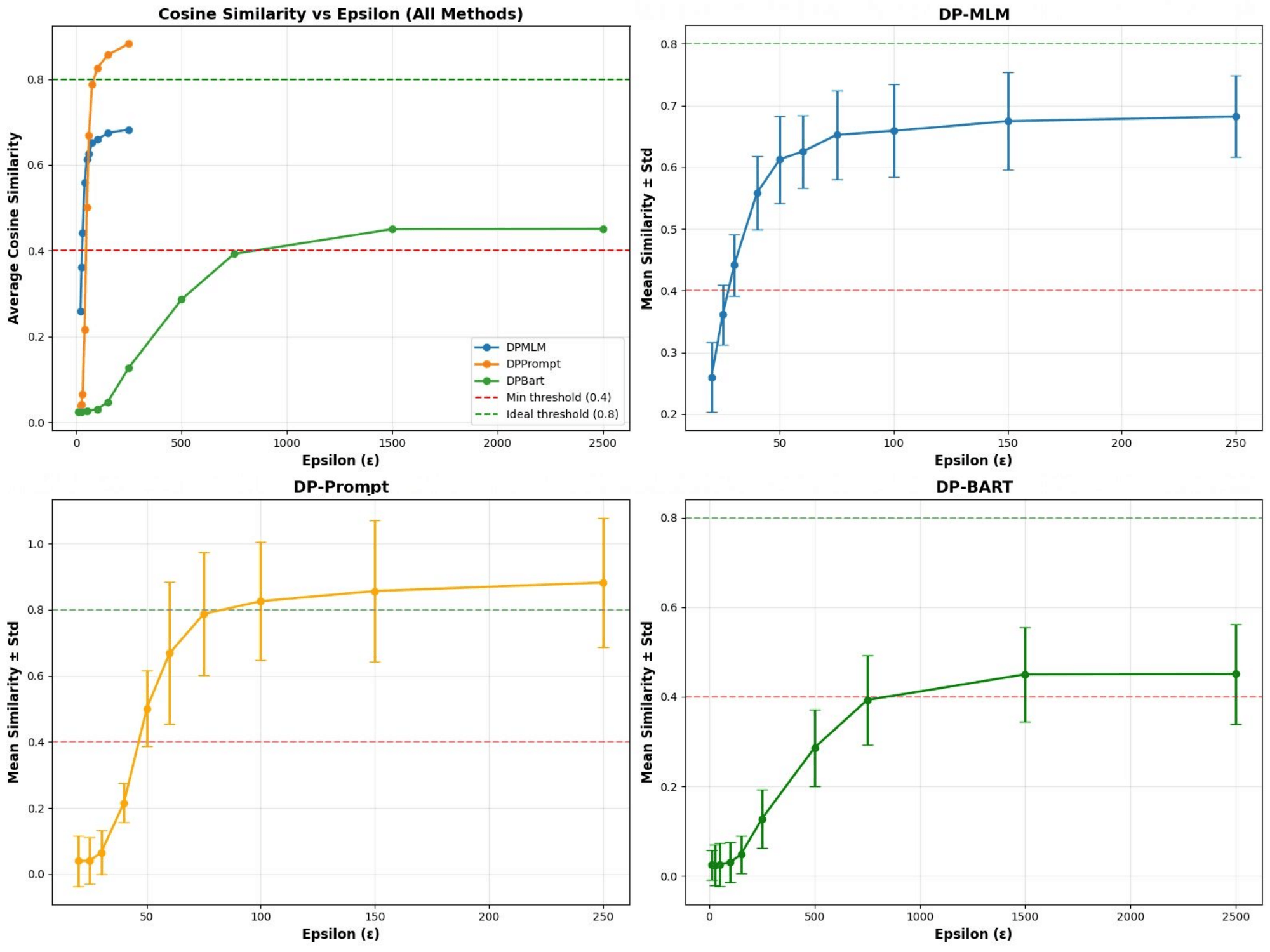}
  \caption{Mean cosine similarity between
    original and privatized texts as a function of $\varepsilon$, computed
    over 20 generations per (method, $\varepsilon$) on our selected text samples. Error bars show $\pm 1$ standard deviation. Dashed lines mark the minimum (\textcolor{BrickRed}{red}, $\mathrm{CS} = 0.4$) and upper target (\textcolor{ForestGreen}{green}, $\mathrm{CS} = 0.8$) thresholds. The top-left panel overlays all three methods on a shared axis; the remaining panels show per-method results.}
\label{fig:cs_vs_epsilon}
\end{figure}

Figure \ref{fig:cs_vs_epsilon} illustrates the results of our calibration tests, which plot privacy budget ($\varepsilon$) against mean cosine similarity (CS) of their privatized outputs. For these tests, we used the \textsc{all-MiniLM-L6-v2} embedding model \cite{reimers-gurevych-2019-sentence}.

In addition, we provide our rationale on the chosen $\varepsilon$ values per mechanisms, in light of the calibration tests and by setting a target maximum threshold of 80\% CS and a minimum of 40\%:

\begin{itemize}[leftmargin=1em]
    \itemsep 0em
    \item \textsc{DP-MLM} ($\varepsilon \in \{25, 50, 100\}$) covers the range from near the minimum CS threshold up to the plateau at roughly CS = 0.7.
    \item \textsc{DP-Prompt} ($\varepsilon \in \{50, 60, 75\}$) is intentionally narrow since the CS curve rises steeply therein. 
    \item \textsc{DP-BART} ($\varepsilon \in \{750, 1500, 2500\}$) spans the region where CS first crosses the minimum threshold and subsequently plateaus. 
\end{itemize}

\subsection{LLM Prompts}
Figure \ref{fig:llm_judge_prompt} provides the prompt used with our five LLM judges for validation of the acceptability metric, namely to complete a mirror image of the human preference survey. Figure \ref{fig:anchor_prompt} describes the prompt used in Configuration 3.1 of our alignment case study, which anchors an LLM privatized text against all DP privatized candidate outputs.

\begin{figure}[!htbp]
\begin{tcolorbox}[
  colback=gray!5,
  colframe=gray!40,
  title={\small LLM Evaluation Prompt},
  fonttitle=\bfseries\small,
  left=2pt, right=2pt, top=2pt, bottom=2pt
]
\footnotesize
You are evaluating rephrased versions of personal text produced by a differential privacy (DP) algorithm. Differential privacy adds mathematical noise to text during processing, which can introduce grammar errors, unnatural phrasing, or loss of meaning. You will see one ORIGINAL sentence and three rephrased options A, B, and C.

\medskip
\textbf{Your task:} choose the single option that is MOST acceptable --- i.e., the one that best preserves the meaning of the original text while being:

\begin{itemize}[leftmargin=*, itemsep=0pt]
  \item most private (i.e., least revealing of sensitive information in your opinion),
  \item easiest to understand (clearest in topic),
  \item most coherent and fluent, and
  \item easiest to read.
\end{itemize}

\smallskip
Reply with ONLY a single letter: A, B, or C. Do NOT explain your choice. Do NOT output anything else.

\smallskip
Please use your own understanding of ``privacy'' --- we do not provide a definition because we want your natural judgment about what feels private.
\end{tcolorbox}
\caption{Zero-shot prompt used for LLM-as-a-Judge selection and acceptability evaluation. The prompt mirrors the instructions of the human preference survey.}
\label{fig:llm_judge_prompt}
\end{figure}

\begin{figure}[htbp]
\begin{tcolorbox}[
  colback=gray!5,
  colframe=gray!40,
  title={\small Anchor Generation Prompt},
  fonttitle=\bfseries\small,
  left=2pt, right=2pt, top=2pt, bottom=2pt
]
\footnotesize
You are an expert privacy-preserving linguist. Your task is to rewrite the provided input text to protect the author's privacy while maximizing the utility and readability of the text. Based on extensive human research, you must adhere to the following rules when rewriting:
 
\begin{enumerate}[leftmargin=*, label=\textbf{\arabic*.}, itemsep=0pt]
 
  \item \textbf{Targeted Masking (Entity-Centric Privacy)}
  \begin{itemize}[leftmargin=*, itemsep=1pt]
    \item Remove or generalize specific personally identifiable information (PII) such as exact names, locations, dates and times, workplaces, and brands.
    \item Do not rewrite the whole sentence if only a specific entity needs masking --- swap only the entity (e.g., \textit{``John went to Starbucks in Seattle''} $\rightarrow$ \textit{``The person went to a coffee shop in the city''}).
  \end{itemize}
 
  \item \textbf{Semantic Preservation (Faithfulness)}
  \begin{itemize}[leftmargin=*, itemsep=1pt]
    \item Preserve the core meaning, narrative, and general factual situation of the original text.
    \item Do not add new information, hallucinate opposite facts, or change the overarching intent.
  \end{itemize}
 
  \item \textbf{Fluency and Coherence (Cognitive Ease)}
  \begin{itemize}[leftmargin=*, itemsep=1pt]
    \item The rewritten text must read naturally. Prioritize smooth grammar and spelling.
    \item Avoid awkward or robotic word substitutions. If a direct substitution sounds unnatural, rephrase the clause so it flows naturally in conversational English.
  \end{itemize}
 
  \item \textbf{Appropriateness and Tone (Sentiment Preservation)}
  \begin{itemize}[leftmargin=*, itemsep=1pt]
    \item Maintain the original emotional tone (e.g., if it is a complaint, keep it a complaint).
    \item If the text contains severe vulgarity or extreme profanity, neutralize it slightly while preserving the underlying emotion.
  \end{itemize}
 
  \item \textbf{Contextual Awareness}
  \begin{itemize}[leftmargin=*, itemsep=1pt]
    \item Only privatize what actually needs to be privatized. If the text contains no sensitive information, make minimal or no changes.
  \end{itemize}
 
\end{enumerate}
 
\smallskip
\textbf{Input:} \texttt{\{original\_text\}}
 
\smallskip
Provide \textbf{only} the final rewritten text.
\end{tcolorbox}
\caption{LLM prompt used to generate the anchor candidate $c_{\text{LLM}}$ based on themes from the thematic analysis.}
\label{fig:anchor_prompt}
\end{figure}

\subsection{Survey Demographics}
\label{sec:demo}
294 participants successfully completed the survey (i.e., passed attention checks). 211 also provided a response to the optional freeform question. The participants spanned all major continents: Europe (n=84), Africa (n=71), North America (n=54), Asia (n=44), South America (23), and Oceania (n=18). The gender distribution was male-skewed (n=176), in comparison to females (n=112) and non-binary (n=6). Participants in the age range of 25-34 (n=125) comprised the largest portion, followed by 35-44 (n=62), 18-24 (n=60), 45-54 (n=31), and 55+ (n=16). The majority had completed a Bachelor's (n=148) or Master's (n=62) degree, along with high school (n=55), Associate's (n=18), or Doctoral (n=11) degrees. Participants reported being employed full-time (n=170) and part-time (n=54), or as a student (n=34), unemployed (n=21), or other (n=15).

\begin{table*}[htbp]
\footnotesize
\centering
\resizebox{\linewidth}{!}{
\begin{tabular}{p{2.5cm}p{2.5cm}p{4.2cm}p{0.3\textwidth}}
\toprule
\textbf{Theme} & \textbf{Sub-theme} & \textbf{Description} & \textbf{Representative quote} \\
\midrule

\multirow{5}{2.8cm}{Heuristics \& Reasoning Patterns}
  & Balancing act
  & Simultaneous optimization of meaning, readability, and privacy with no fixed priority weighting.
  & ``I tried to strike a balance between conveying the original message and making the text feel more anonymous.'' \\
\addlinespace
  & Cognitive ease \& fluency
  & Default to the easiest-to-read option under cognitive load.
  & ``I picked the ones that sounded the most natural and were easy to read.'' \\
\addlinespace
  & Selecting the lesser evil
  & Damage-control heuristic when no viable option exists.
  & ``Most of the time it was just picking the least bad option, honestly.'' \\
\addlinespace
  & Privacy vs.\ utility
  & Explicit directional preference; some humans prioritized coherence over privacy, and others the reverse.
  & ``I often opted for a little less privacy in order to choose the text that maintained meaning a little more.'' \\
\addlinespace
  & Affective/intuitive heuristics
  & Low-effort, gut-feel evaluation by surface similarity or instinct.
  & ``I chose based on the options given and what I felt was right.'' \\
\midrule

\multirow{4}{2.8cm}{User Perception of Privacy}
  & Targeted entity masking
  & Preference for selective removal of PII (names, locations, dates) over global noise.
  & ``I felt texts that removed specific names, places, dates felt more private, while still keeping the general topic recognizable.'' \\
\addlinespace
  & Perceived vulnerabilities
  & Entity-centric model of what constitutes private information.
  & ``I was paying attention to people's or companies' names.'' \\
\addlinespace
  & Unnecessary privatization
  & Source texts perceived as containing no sensitive information.
  & ``To be honest, those texts didn't really have to be privatized.'' \\
\addlinespace
  & Failure in privatization
  & No option perceived as meaningfully more private than the original.
  & ``It didn't really change anything privacy-wise.'' \\
\midrule

\multirow{3}{2.8cm}{Meaning \& Narrative Preservation}
  & Retention of specific details
  & Preservation of key objective information even where general meaning shifts.
  & ``I looked at which did a better job of preserving the largest percentage of key details.'' \\
\addlinespace
  & Faithfulness as constraint
  & Active rejection of options introducing conflicting or fabricated information.
  & ``I was inclined to discount a text if it gave actively conflicting information from the original.'' \\
\addlinespace
  & Meaning over grammar
  & Tolerance for syntactic degradation where core meaning remains recoverable.
  & ``I would look for the same meaning besides the bad English and sentence construction.'' \\
\midrule

\multirow{3}{2.8cm}{Difficulty of Privatized Texts}
  & Linguistic noise
  & Spelling errors and syntactic disruption impairing readability.
  & ``Unintelligible due to many spelling mistakes.'' \\
\addlinespace
  & Severe semantic loss
  & Nonsensical substitutions severing the output from source meaning.
  & ``Some alternatives were too distorted by noise to be considered useful.'' \\
\addlinespace
  & Severe comprehension loss
  & Terminal incoherence rendering meaningful evaluation impossible.
  & ``There were a few questions where I ended up having to choose between three strings of utter nonsense.'' \\
\midrule

\multirow{4}{2.8cm}{Context}
  & Context-dependent privacy tolerance
  & Required privatization degree varies with content sensitivity and communicative stakes.
  & ``Topics that seemed sensitive required a level of detail for readers to provide the requested help.'' \\
\addlinespace
  & Appropriateness \& toxicity avoidance
  & Penalization of options containing vulgar or violent language.
  & ``I didn't choose any that contained vulgar language or simply made no sense.'' \\
\addlinespace
  & Preservation of sentiment \& tone
  & Value placed on emotional register of the original text.
  & ``I chose the one that most closely matched the sentiment of the original, while still making sense.'' \\
\addlinespace
  & Pragmatic failure
  & Privatized text fails to fulfill the communicative function of the original.
  & ``Personally, I wouldn't speak like any of these responses.'' \\

\bottomrule
\end{tabular}
}
\caption{Full codebook with sub-themes, codes, and representative quotes, resulting from our reflexive thematic analysis of freeform survey responses.}
\label{tab:themes_full}
\end{table*}

\begin{table*}[t!]
\centering
\small
\resizebox{\linewidth}{!}{
\begin{tabular}{ccccc}
\toprule
\textbf{ID} & \textbf{LLM Label} & \textbf{$n$} & \textbf{Share}
& \textbf{Top Keywords (c-TF-IDF)} \\
\midrule
0 & Coherence and Meaning Preservation
  & 301 & 65.7\% & original, meaning, options, sense, tried \\
1 & Balancing Privacy with Text Coherence
  & 73  & 15.9\% & private, privacy, information, names, personal \\
2 & Balancing Anonymity with Semantic Preservation
  & 43  & 9.4\%  & personal, details, kept, names, meaning, sensitive \\
3 & Minimizing Sensitive Information Disclosure
  & 41  & 9.0\%  & sensitive, information, specific, preferred, details \\
\bottomrule
\end{tabular}
}
\caption{BERTopic clusters derived from segmented survey feedback sentences.}
\label{tab:bertopic_themes}
\end{table*}

\subsection{Reflexive TA and BERTopic Validation }
\label{sec:bertopic}
Table \ref{tab:themes_full} presents the complete codebook as a result of our reflexive thematic analysis of the 211 freeform feedback responses from the survey.

As mentioned in Section \ref{sec:analysis}, we ran a BERTopic analysis to the same feedback texts as an independent check on the manually derived themes from the reflexive TA. The goal was not to replace this analysis but to assess whether an unsupervised approach would recover a structurally similar thematic codebook.

The feedback responses were first segmented into individual sentences (minimum 12 characters). Sentences were embedded using \textsc{all-MiniLM-L6-v2} \cite{reimers-gurevych-2019-sentence}, reduced to a 5-dimensional space via
UMAP \cite{mcinnes2018umap}, and clustered using HDBSCAN \cite{mcinnes2017hdbscan} with a minimum cluster size of 20. Cluster vocabulary was extracted using c-TF-IDF, and theme labels were generated by prompting \textsc{llama-3.3-70B} \cite{grattafiori2024llama3herdmodels} with the top keywords and representative sentences per cluster.
The resulting clusters and their labels are provided in Table \ref{tab:bertopic_themes}.

\subsection{Alignment Training}
\label{sec:alignment}

\paragraph{Hardware.}
All training was performed on an NVIDIA V100 (16GB VRAM), on a virtual machine with 384GB RAM. The software stack consisted of PyTorch, Hugging Face Transformers, and Transformers Reinforcement Learning (TRL) for the DPO and VPO training objectives.

\paragraph{Model architecture.}
All alignment configurations build on \textsc{facebook/bart-base} \cite{lewis-etal-2020-bart} using the \textsc{DP-BART-CLV} variant of \textsc{DP-BART} \cite{igamberdiev-habernal-2023-dp}, specifically with a clipping range of (-0.1, 0.1). TRL’s \textsc{DPOTrainer} does not natively support DP noise injection during the forward pass, so we wrapped the model in a custom class that injects the necessary calibrated Gaussian noise into the encoder’s hidden states before they reach the decoder. During autoregressive generation, encoder outputs are cached after the first forward pass to prevent repeated noise injection across decoding steps. Both the policy and reference models use this wrapper, so DPO computes log-ratios between two noisy distributions rather than comparing a noisy policy against a clean reference.

As noted in the work, only the decoder underwent alignment training. To implement this, we detached the decoder’s embedding matrix from the shared encoder-decoder embeddings and replaced it with a trainable deep copy. The encoder and shared token embeddings were thus frozen. The language model head is similarly untied and linked to the new decoder embeddings, ensuring that all gradient updates were confined to the decoder.
These architectural and training customizations are included in our public repository.

\paragraph{Privacy budgets.}
As noted in the work, we select three privacy budgets $\varepsilon \in \{250, 750, 1500\}$ for alignment training and evaluation. We train one aligned model for each configuration and $\varepsilon$ pair; thus $9 \times 3 = 27$ aligned models are created. These are all included in our repository.

\paragraph{Training hyperparameters.}
We use the same training hyperparameters across all training runs (Configurations 1-3). These are found in Table \ref{tab:global_hyperparams}.

We set the DPO $\beta$ parameter to 0.05 after initial experiments with $\beta \in {0.01, 0.05, 0.1}$. Given the relatively small preference dataset size, a permissive $\beta$ is necessary to let the policy drift meaningfully from the reference distribution. We found that $\beta$ = 0.01 lead to excessive drift without improving reward margins, while $\beta$ = 0.1 was too conservative to produce meaningful separation.

\begin{table}[H]
\small
\centering
\begin{tabular}{ll}
\toprule
\textbf{Hyperparameter} & \textbf{Value} \\
\midrule
Optimizer                  & AdamW \\
Learning Rate              & 5e-6 \\
LR Schedule                & Cosine decay with warmup \\
Warmup Ratio               & 0.1 \\
Weight Decay               & 0.05 \\
Max Gradient Norm          & 0.5 \\
DPO $\beta$                & 0.05 \\
Max Sequence Length         & 1024 tokens \\
Per-device Batch Size       & 4 \\
Gradient Accumulation       & 8 (effective batch size: 32) \\
Epochs                      & 5 \\
\bottomrule
\end{tabular}
\caption{Global training hyperparameters used across all alignment configurations.}
\label{tab:global_hyperparams}
\end{table}

The only individual deviations to Table \ref{tab:global_hyperparams} come with Configuration 2.3, where the 720-variant is trained for two epochs and the 1800 variant for one epoch, and similarly, for Configuration 3, which is also trained for one epoch. These reductions were undertaken due to the larger training dataset sizes.

\subsection{Evaluation Datasets}
\label{sec:appdatasets}
As mentioned in Section \ref{sec:eval}, we evaluate our aligned models on a subset of the ten datasets introduced in Section \ref{sec:construct}, namely Reddit Mental Health Posts, Trustpilot Reviews, and Yelp Reviews. We note that in the case of all three, the evaluation datasets represent strictly disjoint sets from those used in the preference dataset construction, survey, or alignment training. For each, we took the follow pre-processing measures:

\begin{itemize}
    \itemsep 0em
    \item We use the subset of Reddit Mental Health Posts curated by \citet{10.1145/3733816.3760760}, which confines the data to the top-50 most frequent authors in the dataset. This enables a 50-class authorship identification task. Posts marked as “[removed]” and missing-author entries are excluded.
    \item Trustpilot Reviews are taken from the original corpus \cite{10.1145/2736277.2741141}, and we take a random sample (seed 42) of the \texttt{en-US} split. All reviews are marked with a sentiment (0/1) label, as well as a gender (male/female). Reviews without both of these labels are excluded before sampling.
    \item Yelp Reviews are taken from the sample prepared by \cite{utpala-etal-2023-locally}. This sample is confined to 10 frequently-writing users, thus enabling a 10-class authorship identification task. The reviews are likewise label by sentiment (0/1). Otherwise, we make not modifications to the dataset.
\end{itemize}

Following the above steps, we apply two additional pre-processing measures to all three evaluation datasets. Firstly, we normalize irregular whitespace / newlines and use \textsc{langdetect} to remove an non-English rows. Secondly, we limit the datasets to only those containing 200 words or less (as per \textsc{nltk}), matching the generation parameters used for all \textsc{DP-BART} setups. For Trustpilot, where the corpus is large and nearly all rows already fall within this limit, a further 50\% sample is taken. For
the remaining two datasets, all eligible rows are kept. Table \ref{tab:eval_dataset_stats} summarizes the key evaluation dataset characteristics, both pre- and post-filtering.

\begin{table*}[htbp]
\small
\centering
\begin{tabular}{cccccc}
\hline
\textbf{Dataset} & \textbf{Total Rows} & \textbf{Filtered Rows} & \textbf{Rows Sampled} & \textbf{Avg. Words} & \textbf{Median Words} \\
\hline
Trustpilot              & 29,253 & 28,843 (98.6\%) & 14,422 (50\%) & 48.1  & 39.0  \\
Yelp                    & 17,058 & 11,182 (65.6\%) & 11,182        & 114.6 & 115.0 \\
Reddit-Mental-Health    &  2,068 &  1,664 (80.5\%) &  1,664        & 74.0  & 59.0  \\
\hline
\textbf{Total}          & 48,379 & 41,689          & \textbf{27,268} & -- & -- \\
\hline
\end{tabular}
\caption{Evaluation dataset makeup. Total rows reflect rows of at most 200 words after language filtering.}
\label{tab:eval_dataset_stats}
\end{table*}

\subsection{Evaluation Metrics}
\label{sec:appmetrics}
As introduced in Section \ref{sec:eval}, we employ an array of metrics to measure the utility preservation, privacy protection, and trade-offs afforded by our alignment process. The following describes each metric and the rationale for its inclusion.

\paragraph{Semantic Similarity.}
For each privatized dataset, pairwise cosine similarity is computed between the embeddings of the original text and its privatized counterpart. Both texts are embedded using three \textsc{SentenceTransformer} models: \textsc{all-MiniLM-L12-v2} \cite{reimers-gurevych-2019-sentence}, \textsc{all-mpnet-base-v2} (ibid), and \textsc{gte-small} \cite{li2023generaltextembeddingsmultistage}. The per-pair score is the mean of the three model scores, and the per-dataset score is the mean over all sentence pairs.

\paragraph{Downstream Utility.}
Utility quantifies the extent to which downstream task performance is preserved through text privatization.
We fine-tune a \textsc{DeBERTa-v3-base} \cite{he2021debertadecodingenhancedbertdisentangled} classifier on each privatized dataset for one epoch using a 90/10 train/validation split, with the micro-F1 on the validation split recorded as the utility score. Note that this is only performed for Trustpilot and Yelp, where the downstream task is binary sentiment classification. Per-dataset fine-tuning hyperparameters and target attributes are summarized in Table \ref{tab:finetune_eval_privacy_utility}.

\begin{table*}[htbp]
    \centering
    \small
    \resizebox{\linewidth}{!}{
    \begin{tabular}{cccccc}
    \toprule
    \textbf{Dataset} & \textbf{Target Attributes (Util./Priv.)} & \textbf{Max Input Length} & \textbf{Physical Batch} & \textbf{Accum. Steps} & \textbf{Effective Batch} \\
    \midrule
    Yelp                 & Sentiment/Author ID & 512 & 4 & 8 & 32 \\
    Trustpilot           & Sentiment/Gender    & 256 & 8 & 8 & 64 \\
    Reddit Mental Health & None/Author ID&  256 & 8 & 8 & 64 \\
    \bottomrule
    \end{tabular}
    }
    \caption{Target attributes and fine-tuning hyperparameters per dataset for the downstream utility and empirical privacy evaluations. Note that Reddit does not contain a utility task.}
    \label{tab:finetune_eval_privacy_utility}
\end{table*}

\paragraph{Empirical Privacy.}
While the privacy budget $\varepsilon$ provides a theoretical guarantee in DP text privatization, evaluation should also consider protections against concrete adversaries \cite{mattern-etal-2022-limits,utpala-etal-2023-locally}. For each privatized dataset, a \textsc{DeBERTa-v3-base} is fine-tuned for one epoch to predict the protected attribute, following the same procedure as outlined above for utility measurements. Performance is measured on the 10\% validation split, and empirical privacy is reported as the reduction in F1 relative to a model trained on clean, non-privatized text.

Two adversarial settings are considered, following the previous literature. The \textit{static attacker} (\textit{s}) is trained on clean data and evaluated on the privatized validation split, simulating a scenario where an adversary builds a profiling model from public data before attempting to re-identify users in a sanitized release. The \textit{adaptive attacker} (\textit{a}) trains and evaluates entirely on privatized text, representing a stronger threat model where the adversary knows the privatization algorithm and trains a model on these noisy patterns \cite{mattern-etal-2022-limits}.

The metric we report is the \textit{empirical privacy gain}, defined for each setting as:
\begin{displaymath}
    EP_s = F_1^{base} - F_1^{static}
\end{displaymath}
\begin{displaymath}
    EP_a = F_1^{base} - F_1^{adapt}
\end{displaymath}

where $F_1^{base}$ is the attacker’s micro-F1 on non-privatized text, and $F_1^{static/adapt}$ is the (ideally reduced) performance when trying to adversarial infer from privatized texts. A higher value indicates that DP text privatization left the attacker with less usable signal. When a single score is needed (i.e., for reporting in Figure X), we average the two:

\begin{displaymath}
    EP = \frac{1}{2}(EP_s + EP_a)
\end{displaymath}

\paragraph{G-Eval.}
Coherence, PII leakage, and acceptability are all evaluated using the LLM-as-a-Judge framework \textsc{G-Eval} \cite{liu-etal-2023-g}. Following related work applying \textsc{G-Eval} to natural language generation evaluation \cite{song-etal-2024-finesure,afzal-etal-2024-adapteval}, we use the \textsc{DeepEval} library\footnote{\scriptsize\url{https://deepeval.com/}} to operationalize each criterion. \textsc{DeepEval} takes a one-sentence criterion, converts it into a chain-of-thought evaluation procedure, and returns a score on the [0, 1] scale (higher is better). For all \textsc{G-Eval} evaluations, we use \textsc{gpt-5.4-nano}\footnote{\scriptsize\url{https://openai.com/index/introducing-gpt-5-4/}}.

For \textbf{Coherence}, we measure the mean of five criteria, each targeting a distinct dimension of output quality, following recommendations from the \textsc{DeepEval} documentation:

\begin{enumerate}
    \itemsep 0em
    \item \textbf{Fluency} measures how smoothly the text reads, focusing on grammar and syntactic well-formedness.
    \item \textbf{Consistency} assesses whether the text maintains a uniform style and tone throughout.
    \item \textbf{Clarity} evaluates how easily the output can be understood by the reader.
    \item \textbf{Conciseness} checks whether the text is free of unnecessary words or details.
    \item \textbf{Repetitiveness} identifies redundant or repeated information.
\end{enumerate}

We measure \textbf{PII Leakage} as a complementary privacy measure, i.e, measuring how well personally identifiable information is protected at a surface level. Note that we measure \say{surface} leakage protection, as we do not measure more ingrained identifiers such as indirect personal identifiers \cite{baroud-etal-2025-beyond}. We use the \textsc{DeepEval}'s PII leakage metric\footnote{\scriptsize\url{https://deepeval.com/docs/metrics-pii-leakage}}, which extracts individual factual claims from the text and classifies each as containing PII or not, according to the following:
\begin{displaymath}
    \text{PII Leakage} = \frac{\text{Number of Non-PIIs}}{\text{Total Number of Extracted PIIs}}
\end{displaymath}

The final score therefore represents the proportion of claims classified as safe, so higher values indicate less residual PII.

As the final \textsc{G-Eval} based metric, we measure \textbf{Acceptability} as a reference-free measure independent of our chosen acceptability metrics (Table \ref{tab:chosen-metrics}, \say{Metric Acceptability}). We base our criteria on the uncovered themes from the thematic analysis of the survey, and they are operationalized as such:

\begin{enumerate}
    \itemsep 0em
    \item \textbf{Narrative Faithfulness} requires that the privatized text retain the core narrative intent of the original without introducing conflicting facts or reversing the meaning. Specific names or dates may be legitimately altered, but the overall gist must survive.
    \item \textbf{Comprehensibility} requires that the text be logically understandable. Even with imperfect grammar or simplified vocabulary, the reader should be able to follow it without encountering incoherent words or phrases.
    \item \textbf{Emotional Tone and Appropriateness} requires that the privatized text match the emotional register of the original and not introduce inappropriate, toxic, or vulgar language absent from the source.
    \item \textbf{Targeted Entity Masking} evaluates whether the rewrite successfully obfuscates personal identifiers such as names, exact locations, and brands while keeping the general non-sensitive topic recognizable.
\end{enumerate}

\paragraph{Metric Acceptability.}
Alongside the G-Eval acceptability criteria, we report a Metric Acceptability score. This score is identical to that introduced in Appendix \ref{sec:metrics}, which aggregates three facets covering semantic fidelity, fluency, and readability, and which served as the alignment target during training. Because the evaluation datasets are held out from training, the facet z-scores are recomputed against each evaluation dataset rather than the subsets used during dataset construction and training.

\paragraph{Trade-off and composite scores.}
The first composite score is the \textbf{Composite Utility Ratio}, which is the unweighted mean of our three utility-based measures: downstream task utility (\textit{Util}), coherence (\textsc{G-Eval}, \textit{Coh}), and semantic similarity (\textit{Sim}). This score provides a general understanding of the utility-preserving capabilities of a text privatization method, and it is calculated as follows (higher is better):
\begin{displaymath}
    U_{\text{ratio}} = \frac{1}{3}\left(\frac{F_{1,p}^{\text{util}}}{F_{1,o}^{\text{util}}} + \frac{\text{Coh}_p}{\text{Coh}_o} + \text{Sim}_p\right)
\end{displaymath}

Likewise, a \textbf{Composite Privacy Ratio} is calculated, using both the empirical privacy (EP) and \textsc{G-Eval} PII leakage measurements. As the \textsc{G-Eval} score is reported on a higher-is-safer scale, we first invert it before inclusion in the ratio:
\begin{displaymath}
    L_{\text{ratio}} = \frac{1 - \text{PII}_p}{1-\text{PII}_o}
\end{displaymath}

The composite privacy ratio is thus:
\begin{displaymath}
    P_{\text{ratio}, \alpha} = \frac{1}{2}\left(\frac{F_{1,p}^{\text{EP}}}{F_{1,o}^{\text{EP}}} + L_{\text{ratio}}\right), \ \alpha \in \{s, a\}
\end{displaymath}

From these composite scores follows a measurement of the \textbf{privacy-utility trade-off}, or rather the \textit{relative gain}, of DP text privatization. Intuitively, this measures whether the privacy gains outweigh any utility losses, and thereby signals whether privatization has intrinsic value. Defined by \cite{mattern-etal-2022-limits} and used in the extant literature, relative gain compares retained utility (\textit{U}) against privacy leakage (\textit{P}) for the privatized text (\textit{p}) relative to the baseline, non-privatized original (o):
\begin{displaymath}
    \gamma = \frac{U_p}{U_o} - \frac{P_p}{P_o}
\end{displaymath}

Using our composite score, the relative gain calculation directly follows:
\begin{displaymath}
    \gamma_\alpha = U_{\text{ratio}} - P_{\text{ratio}}, \ \alpha \in \{s, a\}
\end{displaymath}

We note that where a downstream utility task does not exist (here, Reddit), the composite utility ratio simply omits this term and adjust the unweighted mean. We also note that, as displayed above, relative gain is calculated separately for $\gamma_{static}$ and $\gamma_{adapt}$, but these are averaged for the reporting in Figure X, as follows:

\begin{displaymath}
    \gamma = \frac{1}{2}(\gamma_s + \gamma_a)
\end{displaymath}

As the core of our work is centered on the notion of acceptability, we also introduce a second trade-off score, which measures the relative gain between acceptability (from our metrics, not \textsc{G-Eval}) and privacy. This is defined as:
\begin{displaymath}
    \gamma_{\text{acc},\alpha} = \frac{A_p}{A_o} - P_{\text{ratio},\alpha}, \ \alpha \in \{s, a\}
\end{displaymath}

As with $\gamma$, the un-subscripted $\gamma_{\text{acc}}$ denotes the average of the static and adaptive variants.

\subsection{Evaluation Results}
\label{sec:complete}

Tables \ref{tab:reddit_privacy_utility}, \ref{tab:trustpilot_privacy_utility}, and \ref{tab:yelp_privacy_utility} present the  complete privacy and utility results for our three evaluation datasets. \textit{SS} = Semantic Similarity, \textit{Util} = Downstream Utility (micro-$F_1$), \textit{COH} = Coherence, \textit{PII} = PII Leakage ($\uparrow$), \textit{EP}$_s$/\textit{EP}$_a$ = Empirical Privacy Gain (static/adaptive), $\gamma_s$/$\gamma_a$ = Relative Gain (static/adaptive).

Similarly, Tables \ref{tab:reddit_acceptability}, \ref{tab:trustpilot_acceptability}, and \ref{tab:yelp_acceptability} provide the acceptability results. \textit{GE}$_{\text{acc}}$ = \textsc{G-Eval} Acceptability,
\textit{MA} = Metric Acceptability, $\gamma_{\text{acc,s}}$/$\gamma_{\text{acc,a}}$ = Acceptability
Relative Gain (static/adaptive).

Table \ref{tab:app_param_ttests} provides the full statistical test results for the per-configuration t-tests, as introduced in the main work.

\begin{table*}[!hp]
\vfill
\centering
\resizebox{\textwidth}{!}{
\begin{tabular}{l|ccccccc|ccccccc|ccccccc}
\hline
& \multicolumn{7}{c|}{$\epsilon = 250$} & \multicolumn{7}{c|}{$\epsilon = 750$} & \multicolumn{7}{c}{$\epsilon = 1500$} \\
\cline{2-22}
& SS & COH & PII & EP$_s$ & EP$_a$ & $\gamma_s$ & $\gamma_a$ & SS & COH & PII & EP$_s$ & EP$_a$ & $\gamma_s$ & $\gamma_a$ & SS & COH & PII & EP$_s$ & EP$_a$ & $\gamma_s$ & $\gamma_a$ \\
\hline
Baseline & - & 0.393 & 0.412 & 0.105 & 0.105 & - & - & - & 0.393 & 0.412 & 0.105 & 0.105 & - & - & - & 0.393 & 0.412 & 0.105 & 0.105 & - & - \\ \hline
DP-BART (Base) & 0.357 & 0.184 & 0.885 & 0.051 & 0.063 & -0.081 & -0.024 & 0.599 & 0.314 & 0.570 & 0.033 & 0.051 & -0.242 & -0.157 & 0.655 & 0.324 & 0.563 & 0.033 & 0.057 & -0.221 & -0.107 \\
DP-BART (Large) & 0.291 & 0.098 & 0.974 & 0.063 & 0.051 & -0.043 & -0.101 & 0.578 & 0.274 & 0.627 & 0.075 & 0.051 & -0.036 & -0.151 & 0.679 & 0.329 & 0.556 & 0.027 & 0.075 & -0.243 & -0.015 \\ \hline \hline
Config 1.1 & 0.390 & 0.217 & 0.861 & 0.045 & 0.057 & -0.089 & -0.032 & 0.612 & 0.291 & 0.598 & 0.033 & 0.063 & -0.234 & -0.091 & 0.662 & 0.322 & 0.626 & 0.033 & 0.057 & -0.167 & -0.053 \\
Config 1.2 & 0.393 & 0.204 & 0.841 & 0.051 & 0.081 & -0.088 & 0.055 & 0.622 & 0.312 & 0.566 & 0.033 & 0.057 & -0.240 & -0.125 & 0.668 & 0.300 & 0.521 & 0.033 & 0.057 & -0.272 & -0.158 \\
Config 2.1 & 0.415 & 0.258 & 0.884 & 0.039 & 0.075 & -0.056 & 0.115 & 0.636 & 0.342 & 0.475 & 0.033 & 0.057 & -0.287 & -0.173 & 0.674 & 0.322 & 0.508 & 0.033 & 0.069 & -0.263 & -0.092 \\
Config 2.2 (Relaxed) & 0.425 & 0.249 & 0.891 & 0.045 & 0.075 & -0.025 & 0.118 & 0.627 & 0.331 & 0.503 & 0.033 & 0.075 & -0.275 & -0.075 & 0.673 & 0.312 & 0.545 & 0.033 & 0.057 & -0.241 & -0.127 \\
Config 2.2 (Strict) & 0.413 & 0.246 & 0.837 & 0.057 & 0.087 & -0.021 & 0.122 & 0.631 & 0.319 & 0.573 & 0.033 & 0.075 & -0.225 & -0.026 & 0.677 & 0.324 & 0.469 & 0.027 & 0.075 & -0.322 & -0.094 \\
Config 2.3 (120) & 0.394 & 0.207 & 0.963 & 0.039 & 0.069 & -0.039 & 0.104 & 0.615 & 0.314 & 0.629 & 0.033 & 0.069 & -0.187 & -0.016 & 0.661 & 0.314 & 0.600 & 0.033 & 0.063 & -0.197 & -0.054 \\
Config 2.3 (720) & 0.415 & 0.211 & 0.906 & 0.045 & 0.075 & -0.048 & 0.094 & 0.637 & 0.305 & 0.571 & 0.027 & 0.075 & -0.265 & -0.037 & 0.692 & 0.306 & 0.551 & 0.033 & 0.057 & -0.234 & -0.120 \\
Config 2.3 (1800) & 0.418 & 0.228 & 0.897 & 0.033 & 0.075 & -0.097 & 0.103 & 0.645 & 0.296 & 0.489 & 0.033 & 0.075 & -0.312 & -0.112 & 0.695 & 0.316 & 0.517 & 0.033 & 0.075 & -0.254 & -0.054 \\
Config 3.1 & 0.455 & 0.237 & 0.594 & 0.039 & 0.069 & -0.306 & -0.164 & 0.668 & 0.308 & 0.396 & 0.027 & 0.051 & -0.401 & -0.287 & 0.708 & 0.278 & 0.394 & 0.033 & 0.069 & -0.386 & -0.215 \\
\hline
\end{tabular}
}
\caption{Privacy and utility results for the Reddit Mental Health dataset. For Tables \ref{tab:reddit_privacy_utility}-\ref{tab:yelp_privacy_utility}, \textit{SS} denotes semantic similarity, \textit{COH} denotes G-Eval coherence, \textit{PII} denotes G-Eval PII leakage, \textit{EP} refers to empirical privacy, and $\gamma$ is the trade-off (relative) gain calculation. In all cases except for \textit{EP}, a higher score is better.}
\label{tab:reddit_privacy_utility}
\end{table*}

\begin{table*}[ht]
\centering
\resizebox{\textwidth}{!}{
\begin{tabular}{l|cccccccc|cccccccc|cccccccc}
\hline
& \multicolumn{8}{c|}{$\epsilon = 250$} & \multicolumn{8}{c|}{$\epsilon = 750$} & \multicolumn{8}{c}{$\epsilon = 1500$} \\
\cline{2-25}
& SS & Util ($F_1$) & COH & PII & EP$_s$ & EP$_a$ & $\gamma_s$ & $\gamma_a$ & SS & Util ($F_1$) & COH & PII & EP$_s$ & EP$_a$ & $\gamma_s$ & $\gamma_a$ & SS & Util ($F_1$) & COH & PII & EP$_s$ & EP$_a$ & $\gamma_s$ & $\gamma_a$ \\
\hline
Baseline & - & 99.72 & 0.432 & 0.833 & 0.710 & 0.710 & - & - & - & 99.72 & 0.432 & 0.833 & 0.710 & 0.710 & - & - & - & 99.72 & 0.432 & 0.833 & 0.710 & 0.710 & - & - \\
\hline
DP-BART (Base) & 0.345 & 94.36 & 0.162 & 0.837 & 0.129 & 0.123 & -0.658 & -0.662 & 0.590 & 94.84 & 0.348 & 0.832 & 0.110 & 0.119 & -0.460 & -0.453 & 0.653 & 98.54 & 0.384 & 0.901 & 0.092 & 0.086 & -0.217 & -0.222 \\
DP-BART (Large) & 0.290 & 93.00 & 0.108 & 0.970 & 0.140 & 0.140 & -0.311 & -0.311 & 0.579 & 99.31 & 0.301 & 0.928 & 0.097 & 0.082 & -0.222 & -0.233 & 0.696 & 99.72 & 0.358 & 0.889 & 0.106 & 0.067 & -0.249 & -0.277 \\
\hline\hline
Config 1.1 & 0.382 & 93.27 & 0.222 & 0.919 & 0.135 & 0.135 & -0.349 & -0.348 & 0.621 & 0.982 & 0.391 & 0.896 & 0.150 & 0.098 & -0.197 & -0.233 & 0.674 & 99.10 & 0.373 & 0.921 & 0.128 & 0.077 & -0.129 & -0.165 \\
Config 1.2 & 0.385 & 92.79 & 0.221 & 0.930 & 0.144 & 0.138 & -0.310 & -0.314 & 0.627 & 98.34 & 0.360 & 0.952 & 0.135 & 0.099 & -0.062 & -0.087 & 0.679 & 98.96 & 0.397 & 0.972 & 0.109 & 0.074 & 0.026 & 0.001 \\
Config 2.1 & 0.398 & 93.14 & 0.238 & 0.932 & 0.182 & 0.154 & -0.259 & -0.279 & 0.640 & 99.31 & 0.359 & 0.933 & 0.151 & 0.083 & -0.104 & -0.152 & 0.690 & 99.31 & 0.402 & 0.902 & 0.119 & 0.071 & -0.169 & -0.203 \\
Config 2.2 (Relaxed) & 0.405 & 93.00 & 0.263 & 0.931 & 0.202 & 0.154 & -0.227 & -0.260 & 0.632 & 98.61 & 0.351 & 0.884 & 0.163 & 0.081 & -0.251 & -0.309 & 0.680 & 99.10 & 0.432 & 0.909 & 0.148 & 0.065 & -0.108 & -0.166 \\
Config 2.2 (Strict) & 0.390 & 93.21 & 0.233 & 0.938 & 0.171 & 0.154 & -0.255 & -0.267 & 0.636 & 99.17 & 0.351 & 0.901 & 0.145 & 0.085 & -0.212 & -0.254 & 0.691 & 99.10 & 0.381 & 0.878 & 0.130 & 0.066 & -0.250 & -0.295 \\
Config 2.3 (120) & 0.378 & 93.42 & 0.241 & 0.956 & 0.144 & 0.154 & -0.218 & -0.211 & 0.618 & 99.38 & 0.350 & 0.949 & 0.153 & 0.103 & -0.069 & -0.104 & 0.674 & 99.10 & 0.400 & 0.883 & 0.128 & 0.080 & -0.227 & -0.261 \\
Config 2.3 (720) & 0.401 & 93.00 & 0.239 & 0.900 & 0.168 & 0.154 & -0.363 & -0.373 & 0.656 & 99.38 & 0.374 & 0.982 & 0.124 & 0.103 & 0.040 & 0.025 & 0.716 & 99.31 & 0.373 & 0.932 & 0.122 & 0.084 & -0.091 & -0.118 \\
Config 2.3 (1800) & 0.406 & 93.42 & 0.273 & 0.952 & 0.179 & 0.154 & -0.172 & -0.189 & 0.661 & 99.38 & 0.362 & 0.933 & 0.130 & 0.098 & -0.110 & -0.132 & 0.721 & 99.38 & 0.382 & 0.926 & 0.107 & 0.080 & -0.111 & -0.130 \\
Config 3.1 & 0.420 & 93.76 & 0.202 & 0.816 & 0.173 & 0.154 & -0.634 & -0.647 & 0.661 & 0.993 & 0.340 & 0.918 & 0.110 & 0.084 & -0.186 & -0.204 & 0.704 & 99.45 & 0.343 & 0.891 & 0.098 & 0.081 & -0.258 & -0.270 \\
\hline
\end{tabular}
}
\caption{Privacy and utility results for the Trustpilot dataset.}
\label{tab:trustpilot_privacy_utility}
\end{table*}

\begin{table*}[ht]
\centering
\resizebox{\textwidth}{!}{
\begin{tabular}{l|cccccccc|cccccccc|cccccccc}
\hline
& \multicolumn{8}{c|}{$\epsilon = 250$} & \multicolumn{8}{c|}{$\epsilon = 750$} & \multicolumn{8}{c}{$\epsilon = 1500$} \\
\cline{2-25}
& SS & Util ($F_1$) & COH & PII & EP$_s$ & EP$_a$ & $\gamma_s$ & $\gamma_a$ & SS & Util ($F_1$) & COH & PII & EP$_s$ & EP$_a$ & $\gamma_s$ & $\gamma_a$ & SS & Util ($F_1$) & COH & PII & EP$_s$ & EP$_a$ & $\gamma_s$ & $\gamma_a$ \\
\hline
Baseline & - & 95.62 & 0.451 & 0.756 & 0.904 & 0.904 & - & - & - & 95.62 & 0.451 & 0.756 & 0.904 & 0.904 & - & - & - & 95.62 & 0.451 & 0.756 & 0.904 & 0.904 & - & - \\
\hline
DP-BART (Base) & 0.335 & 92.00 & 0.186 & 0.856 & 0.652 & 0.603 & -0.186 & -0.213 & 0.547 & 93.03 & 0.297 & 0.901 & 0.635 & 0.399 & 0.050 & -0.080 & 0.607 & 94.59 & 0.358 & 0.871 & 0.547 & 0.158 & 0.004 & -0.211 \\
DP-BART (Large) & 0.280 & 92.85 & 0.085 & 0.977 & 0.696 & 0.626 & -0.006 & -0.045 & 0.550 & 91.83 & 0.315 & 0.827 & 0.671 & 0.347 & -0.068 & -0.247 & 0.666 & 94.01 & 0.362 & 0.849 & 0.588 & 0.314 & 0.005 & -0.146 \\
\hline\hline
Config 1.1 & 0.372 & 92.05 & 0.252 & 0.965 & 0.714 & 0.591 & 0.133 & 0.065 & 0.596 & 93.03 & 0.383 & 0.972 & 0.690 & 0.433 & 0.306 & 0.164 & 0.642 & 94.01 & 0.373 & 0.937 & 0.660 & 0.346 & 0.226 & 0.052 \\
Config 1.2 & 0.377 & 93.03 & 0.229 & 0.975 & 0.718 & 0.568 & 0.141 & 0.058 & 0.601 & 93.03 & 0.350 & 0.925 & 0.685 & 0.398 & 0.184 & 0.025 & 0.646 & 93.21 & 0.355 & 0.934 & 0.667 & 0.340 & 0.211 & 0.030 \\
Config 2.1 & 0.400 & 93.03 & 0.232 & 0.905 & 0.735 & 0.587 & 0.016 & -0.066 & 0.600 & 93.03 & 0.391 & 0.894 & 0.643 & 0.400 & 0.127 & -0.008 & 0.640 & 93.30 & 0.404 & 0.904 & 0.633 & 0.336 & 0.165 & 0.001 \\
Config 2.2 (Relaxed) & 0.418 & 93.03 & 0.256 & 0.925 & 0.743 & 0.566 & 0.086 & -0.012 & 0.600 & 93.03 & 0.351 & 0.935 & 0.664 & 0.424 & 0.193 & 0.060 & 0.640 & 94.01 & 0.375 & 0.913 & 0.629 & 0.358 & 0.160 & 0.010 \\
Config 2.2 (Strict) & 0.393 & 92.94 & 0.268 & 0.929 & 0.747 & 0.585 & 0.096 & 0.007 & 0.604 & 93.12 & 0.355 & 0.884 & 0.680 & 0.413 & 0.102 & -0.046 & 0.648 & 93.92 & 0.404 & 0.886 & 0.643 & 0.333 & 0.136 & -0.035 \\
Config 2.3 (120) & 0.366 & 93.03 & 0.221 & 0.942 & 0.740 & 0.564 & 0.076 & -0.021 & 0.591 & 93.57 & 0.359 & 0.945 & 0.706 & 0.410 & 0.240 & 0.076 & 0.638 & 94.46 & 0.372 & 0.981 & 0.675 & 0.382 & 0.321 & 0.159 \\
Config 2.3 (720) & 0.396 & 93.03 & 0.240 & 0.955 & 0.738 & 0.565 & 0.126 & 0.030 & 0.622 & 94.01 & 0.350 & 0.910 & 0.672 & 0.392 & 0.153 & -0.002 & 0.669 & 93.03 & 0.416 & 0.911 & 0.615 & 0.328 & 0.188 & 0.029 \\
Config 2.3 (1800) & 0.392 & 93.03 & 0.243 & 0.992 & 0.761 & 0.558 & 0.215 & 0.103 & 0.621 & 93.92 & 0.360 & 0.957 & 0.661 & 0.386 & 0.250 & 0.047 & 0.671 & 93.57 & 0.370 & 0.921 & 0.609 & 0.365 & 0.172 & 0.037 \\
Config 3.1 & 0.424 & 93.03 & 0.241 & 0.904 & 0.716 & 0.568 & 0.019 & -0.063 & 0.614 & 0.9392 & 0.335 & 0.856 & 0.727 & 0.432 & 0.059 & -0.104 & 0.653 & 94.82 & 0.346 & 0.825 & 0.702 & 0.364 & 0.003 & -0.184 \\
\hline
\end{tabular}
}
\caption{Privacy and utility results for the Yelp dataset.}
\label{tab:yelp_privacy_utility}
\end{table*}

\begin{table*}[ht]
\centering
\resizebox{0.9\textwidth}{!}{
\begin{tabular}{l|cccc|cccc|cccc}
\toprule
& \multicolumn{4}{c|}{$\epsilon = 250$} & \multicolumn{4}{c|}{$\epsilon = 750$} & \multicolumn{4}{c}{$\epsilon = 1500$} \\
\cline{2-13}
& $GE_{\text{acc}}$ & MA & $\gamma_{acc,s}$ & $\gamma_{acc,a}$ & $GE_{\text{acc}}$ & MA & $\gamma_{acc,s}$ & $\gamma_{acc,a}$ & $GE_{\text{acc}}$ & MA & $\gamma_{acc,s}$ & $\gamma_{acc,a}$ \\
\hline
Baseline & 0.794 & -0.189 & - & - & 0.794 & -0.189 & - & - & 0.794 & -0.189 & - & - \\
\hline
DP-BART (Base) & 0.095 & -0.984 & 0.124 & 0.239 & 0.341 & -0.778 & 0.022 & 0.193 & 0.420 & -0.705 & 0.045 & 0.273 \\
DP-BART (Large) & 0.037 & -1.238 & 0.159 & 0.044 & 0.325 & -0.778 & 0.420 & 0.190 & 0.445 & -0.661 & 0.002 & 0.459 \\
\hline
Config 1.1 & 0.140 & -0.923 & 0.090 & 0.204 & 0.394 & -0.761 & 0.028 & 0.313 & 0.451 & -0.670 & 0.056 & 0.284 \\
Config 1.2 & 0.134 & -0.848 & 0.171 & 0.456 & 0.409 & -0.715 & 0.042 & 0.270 & 0.429 & -0.641 & 0.066 & 0.294 \\
Config 2.1 & 0.194 & -1.207 & -0.058 & 0.285 & 0.450 & -0.829 & 0.006 & 0.234 & 0.500 & -0.737 & 0.035 & 0.377 \\
Config 2.2 (Relaxed) & 0.198 & -1.234 & -0.009 & 0.276 & 0.428 & -0.862 & -0.005 & 0.395 & 0.477 & -0.771 & 0.024 & 0.252 \\
Config 2.2 (Strict) & 0.163 & -1.116 & 0.143 & 0.428 & 0.420 & -0.776 & 0.023 & 0.422 & 0.460 & -0.689 & -0.006 & 0.450 \\
Config 2.3 (120) & 0.128 & -0.914 & 0.036 & 0.321 & 0.435 & -0.750 & 0.031 & 0.373 & 0.458 & -0.669 & 0.057 & 0.342 \\
Config 2.3 (720) & 0.167 & -0.911 & 0.094 & 0.379 & 0.440 & -0.672 & -0.001 & 0.455 & 0.474 & -0.568 & 0.089 & 0.317 \\
Config 2.3 (1800) & 0.174 & -0.940 & -0.030 & 0.370 & 0.446 & -0.665 & 0.058 & 0.457 & 0.469 & -0.566 & 0.090 & 0.489 \\
Config 3.1 & 0.179 & -0.745 & 0.089 & 0.375 & 0.395 & -0.505 & 0.052 & 0.280 & 0.431 & -0.456 & 0.125 & 0.467 \\
\bottomrule
\end{tabular}
}
\caption{Acceptability results for the Reddit Mental Health dataset. In Tables \ref{tab:reddit_acceptability}-\ref{tab:yelp_acceptability}, \textit{GE} denotes G-Eval acceptability, \textit{MA} denotes metric acceptability, and $\gamma$ refers to the acceptability-privacy trade-off metric. In all cases, a higher score is better.}
\label{tab:reddit_acceptability}
\end{table*}

\begin{table*}[th]
\centering
\resizebox{0.9\textwidth}{!}{
\begin{tabular}{l|cccc|cccc|cccc}
\hline
& \multicolumn{4}{c|}{$\epsilon = 250$} & \multicolumn{4}{c|}{$\epsilon = 750$} & \multicolumn{4}{c}{$\epsilon = 1500$} \\
\cline{2-13}
& $GE_{\text{acc}}$ & MA & $\gamma_{acc,s}$ & $\gamma_{acc,a}$ & $GE_{\text{acc}}$ & MA & $\gamma_{acc,s}$ & $\gamma_{acc,a}$ & $GE_{\text{acc}}$ & MA & $\gamma_{acc,s}$ & $\gamma_{acc,a}$ \\
\hline
Baseline & 0.755 & -0.239 & - & - & 0.755 & -0.239 & - & - & 0.755 & -0.239 & - & - \\
\hline
DP-BART (Base) & 0.074 & -1.262 & -0.257 & -0.266 & 0.403 & -0.950 & -0.183 & -0.170 & 0.461 & -0.812 & -0.163 & -0.172 \\
DP-BART (Large) & 0.038 & -1.451 & -0.302 & -0.302 & 0.373 & -0.962 & -0.205 & -0.226 & 0.509 & -0.736 & -0.119 & -0.173 \\
\hline
Config 1.1 & 0.123 & -1.197 & -0.227 & -0.226 & 0.459 & -0.883 & -0.105 & -0.177 & 0.513 & -0.745 & -0.092 & -0.163 \\
Config 1.2 & 0.106 & -1.130 & -0.194 & -0.201 & 0.413 & -0.836 & -0.111 & -0.161 & 0.492 & -0.712 & -0.107 & -0.156 \\
Config 2.1 & 0.160 & -1.467 & -0.248 & -0.288 & 0.459 & -0.915 & -0.113 & -0.209 & 0.535 & -0.770 & -0.111 & -0.179 \\
Config 2.2 (Relaxed) & 0.176 & -1.509 & -0.234 & -0.301 & 0.431 & -0.994 & -0.122 & -0.238 & 0.527 & -0.864 & -0.102 & -0.218 \\
Config 2.2 (Strict) & 0.145 & -1.418 & -0.248 & -0.272 & 0.421 & -0.907 & -0.119 & -0.203 & 0.492 & -0.761 & -0.094 & -0.184 \\
Config 2.3 (120) & 0.117 & -1.239 & -0.228 & -0.214 & 0.456 & -0.893 & -0.103 & -0.175 & 0.537 & -0.751 & -0.092 & -0.160 \\
Config 2.3 (720) & 0.159 & -1.211 & -0.186 & -0.205 & 0.496 & -0.733 & -0.093 & -0.123 & 0.541 & -0.581 & -0.046 & -0.100 \\
Config 2.3 (1800) & 0.177 & -1.255 & -0.184 & -0.219 & 0.485 & -0.736 & -0.086 & -0.130 & 0.554 & -0.573 & -0.065 & -0.102 \\
Config 3.1 & 0.117 & -0.946 & -0.093 & -0.119 & 0.387 & -0.564 & -0.058 & -0.095 & 0.447 & -0.492 & -0.051 & -0.075 \\
\hline
\end{tabular}
}
\caption{Acceptability results for the Trustpilot dataset.}
\label{tab:trustpilot_acceptability}
\end{table*}

\begin{table*}[!ht]
\centering
\resizebox{0.9\textwidth}{!}{
\begin{tabular}{l|cccc|cccc|cccc}
\hline
& \multicolumn{4}{c|}{$\epsilon = 250$} & \multicolumn{4}{c|}{$\epsilon = 750$} & \multicolumn{4}{c}{$\epsilon = 1500$} \\
\cline{2-13}
& $GE_{\text{acc}}$ & MA & $\gamma_{acc,s}$ & $\gamma_{acc,a}$ & $GE_{\text{acc}}$ & MA & $\gamma_{acc,s}$ & $\gamma_{acc,a}$ & $GE_{\text{acc}}$ & MA & $\gamma_{acc,s}$ & $\gamma_{acc,a}$ \\
\hline
Baseline & 0.760 & -0.251 & - & - & 0.760 & -0.251 & - & - & 0.760 & -0.251 & - & - \\
\hline
DP-BART (Base) & 0.044 & -1.111 & 0.334 & 0.280 & 0.180 & -1.128 & 0.310 & 0.049 & 0.291 & -1.654 & 0.041 & -0.389 \\
DP-BART (Large) & 0.026 & -1.188 & 0.358 & 0.280 & 0.215 & -1.481 & 0.235 & -0.124 & 0.348 & -0.994 & 0.301 & -0.001 \\
\hline
Config 1.1 & 0.067 & -1.141 & 0.393 & 0.256 & 0.288 & -1.177 & 0.355 & 0.071 & 0.324 & -1.092 & 0.349 & 0.001 \\
Config 1.2 & 0.062 & -1.085 & 0.415 & 0.250 & 0.266 & -1.143 & 0.361 & 0.042 & 0.313 & -1.060 & 0.367 & 0.005 \\
Config 2.1 & 0.106 & -1.370 & 0.341 & 0.178 & 0.280 & -1.305 & 0.261 & -0.007 & 0.331 & -1.243 & 0.270 & -0.058 \\
Config 2.2 (Relaxed) & 0.121 & -1.407 & 0.338 & 0.142 & 0.292 & -1.328 & 0.277 & 0.011 & 0.329 & -1.274 & 0.257 & -0.044 \\
Config 2.2 (Strict) & 0.102 & -1.324 & 0.370 & 0.191 & 0.287 & -1.232 & 0.326 & 0.030 & 0.331 & -1.155 & 0.309 & -0.033 \\
Config 2.3 (120) & 0.076 & -1.188 & 0.406 & 0.211 & 0.272 & -1.203 & 0.364 & 0.036 & 0.336 & -1.119 & 0.356 & 0.032 \\
Config 2.3 (720) & 0.093 & -1.211 & 0.397 & 0.205 & 0.302 & -1.039 & 0.379 & 0.070 & 0.350 & -0.914 & 0.357 & 0.039 \\
Config 2.3 (1800) & 0.097 & -1.261 & 0.406 & 0.182 & 0.295 & -1.090 & 0.351 & 0.047 & 0.350 & -0.950 & 0.338 & 0.069 \\
Config 3.1 & 0.075 & -0.838 & 0.493 & 0.330 & 0.224 & -0.642 & 0.568 & 0.242 & 0.269 & -0.589 & 0.559 & 0.184 \\
\hline
\end{tabular}
}
\caption{Acceptability results for the Yelp dataset.}
\label{tab:yelp_acceptability}
\end{table*}

\subsection{Output Examples}
\label{sec:examples}
Table \ref{tab:example_detailed} provides a detailed comparison between an unaligned \textsc{DP-BART} output and that from one of our alignment configurations.

Detailed examples from all baselines and alignment configurations can be found in Tables \ref{tab:examples_reddit_mental_health}, \ref{tab:examples_trustpilot}, and \ref{tab:examples_yelp}. Note that these examples were truncated to the input text token length, for readability.

\begin{table}[H]
\centering
\footnotesize
\renewcommand{\arraystretch}{1.3}
\resizebox{\linewidth}{!}{
\begin{tabular}{cp{0.9\linewidth}}
\toprule
\multicolumn{2}{c}{\hfill $\varepsilon = 250$ \quad Dataset: Reddit Mental Health} \\
\midrule
\textbf{Original} &
\textit{``I'm having a hard time with OCD. I feel like I want to die. I don't know how to rely on others to tell me well. I have OCD and intellectual disability as well.''} \\[6pt]

\textbf{DP-BART (base)} &
\textit{``I'm a bit of a talk - and - talk - - - and have a lot of talk. I'm a lot. I have a good time. I don't know how to talk.''} \\[6pt]

\textbf{Config 2.3 (1800)} &
\textit{``I'm having a difficult time with my own experience. I'm having to help with my work. I feel like I'm going to have to help. I think I'm really struggling with my feelings.''} \\
\bottomrule
\end{tabular}
}

\vspace{4pt}
\begin{tabular}{ccccc}
\toprule
& \textbf{SS} & \textbf{COH} & \textbf{MA} & \textbf{ACC} \\
\midrule
Original                 & 1.000 & 0.420 & -0.382 & 0.425 \\
DP-BART (base)           & 0.394 & 0.100 & -0.828 & 0.100 \\
Config 2.3 (1800) & 0.466 & 0.340 & -0.542 & 0.900 \\
\bottomrule
\end{tabular}
\caption{Example outputs for Reddit Mental Health at $\varepsilon = 250$. In order, metrics presented are semantic similarity, coherence (G-Eval), metric acceptability, and G-Eval acceptability.}
\label{tab:example_detailed}
\end{table}

\begin{table}[H]
\centering
\scriptsize
\setlength{\tabcolsep}{3.5pt}

\begin{minipage}[t]{0.48\textwidth}
\centering
\textbf{Global (all $\varepsilon$ pooled)}\vspace{4pt}\\
\begin{tabular}{l l r r r r l}
\hline
\textbf{Group} & \textbf{Metric} & \textbf{T-Mean} & \textbf{B-Mean} & \textbf{p} & \textbf{d} & \textbf{Sig.} \\
\hline
Aligned & Utility  & 0.8991 & 0.8909 & 0.001 & 0.027 & *** \\
Aligned & Cosine   & 0.5649 & 0.4554 & $<$0.001 & 0.647 & *** \\
Aligned & Met.~Acc.& -1.0005 & -1.3366 & $<$0.001 & 0.754 & *** \\
C1.x    & Utility  & 0.8960 & 0.8909 & 0.118 & 0.016 & n.s. \\
C1.x    & Cosine   & 0.5520 & 0.4554 & $<$0.001 & 0.549 & *** \\
C1.x    & Met.~Acc.& -0.9892 & -1.3366 & $<$0.001 & 0.714 & *** \\
C2.x    & Utility  & 0.8995 & 0.8909 & $<$0.001 & 0.028 & *** \\
C2.x    & Cosine   & 0.5662 & 0.4554 & $<$0.001 & 0.646 & *** \\
C2.x    & Met.~Acc.& -1.0593 & -1.3366 & $<$0.001 & 0.603 & *** \\
C3   & Utility  & 0.9031 & 0.8909 & 0.002 & 0.040 & ** \\
C3    & Cosine   & 0.5829 & 0.4554 & $<$0.001 & 0.713 & *** \\
C3    & Met.~Acc.& -0.6705 & -1.3366 & $<$0.001 & 1.303 & *** \\
\hline
\end{tabular}
\end{minipage}\hfill

\vspace{1.5em}

\begin{minipage}[t]{0.48\textwidth}
\centering
\textbf{$\varepsilon = 250$}\vspace{4pt}\\
\begin{tabular}{l l r r r r l}
\hline
\textbf{Group} & \textbf{Metric} & \textbf{T-Mean} & \textbf{B-Mean} & \textbf{p} & \textbf{d} & \textbf{Sig.} \\
\hline
Aligned & Utility  & 0.8759 & 0.8761 & 0.952 & -0.001 & n.s. \\
Aligned & Cosine   & 0.3960 & 0.3101 & $<$0.001 & 1.114 & *** \\
Aligned & Met.~Acc.& -1.2215 & -1.5080 & $<$0.001 & 0.763 & *** \\
C1.x    & Utility  & 0.8728 & 0.8761 & 0.566 & -0.010 & n.s. \\
C1.x    & Cosine   & 0.3804 & 0.3101 & $<$0.001 & 1.049 & *** \\
C1.x    & Met.~Acc.& -1.1258 & -1.5080 & $<$0.001 & 0.893 & *** \\
C2.x    & Utility  & 0.8763 & 0.8761 & 0.968 & 0.001 & n.s. \\
C2.x    & Cosine   & 0.3965 & 0.3101 & $<$0.001 & 1.162 & *** \\
C2.x    & Met.~Acc.& -1.3087 & -1.5080 & $<$0.001 & 0.541 & *** \\
C3    & Utility  & 0.8794 & 0.8761 & 0.650 & 0.010 & n.s. \\
C3    & Cosine   & 0.4238 & 0.3101 & $<$0.001 & 1.661 & *** \\
C3    & Met.~Acc.& -0.8894 & -1.5080 & $<$0.001 & 1.355 & *** \\
\hline
\end{tabular}
\end{minipage}

\vspace{1.5em}

\begin{minipage}[t]{0.48\textwidth}
\centering
\textbf{$\varepsilon = 750$}\vspace{4pt}\\
\begin{tabular}{l l r r r r l}
\hline
\textbf{Group} & \textbf{Metric} & \textbf{T-Mean} & \textbf{B-Mean} & \textbf{p} & \textbf{d} & \textbf{Sig.} \\
\hline
Aligned & Utility  & 0.9095 & 0.8948 & $<$0.001 & 0.050 & *** \\
Aligned & Cosine   & 0.6248 & 0.4781 & $<$0.001 & 1.088 & *** \\
Aligned & Met.~Acc.& -0.9457 & -1.3936 & $<$0.001 & 1.027 & *** \\
C1.x    & Utility  & 0.9050 & 0.8948 & 0.054 & 0.034 & n.s. \\
C1.x    & Cosine   & 0.6132 & 0.4781 & $<$0.001 & 0.907 & *** \\
C1.x    & Met.~Acc.& -0.9750 & -1.3936 & $<$0.001 & 0.851 & *** \\
C2.x    & Utility  & 0.9105 & 0.8948 & $<$0.001 & 0.054 & *** \\
C2.x    & Cosine   & 0.6259 & 0.4781 & $<$0.001 & 1.061 & *** \\
C2.x    & Met.~Acc.& -0.9947 & -1.3936 & $<$0.001 & 0.884 & *** \\
C3    & Utility  & 0.9121 & 0.8948 & 0.010 & 0.057 & ** \\
C3    & Cosine   & 0.6418 & 0.4781 & $<$0.001 & 1.040 & *** \\
C3    & Met.~Acc.& -0.5928 & -1.3936 & $<$0.001 & 1.573 & *** \\
\hline
\end{tabular}
\end{minipage}\hfill

\vspace{1.5em}

\begin{minipage}[t]{0.48\textwidth}
\centering
\textbf{$\varepsilon = 1500$}\vspace{4pt}\\
\begin{tabular}{l l r r r r l}
\hline
\textbf{Group} & \textbf{Metric} & \textbf{T-Mean} & \textbf{B-Mean} & \textbf{p} & \textbf{d} & \textbf{Sig.} \\
\hline
Aligned & Utility  & 0.9119 & 0.9098 & 0.632 & 0.008 & n.s. \\
Aligned & Cosine   & 0.6738 & 0.6653 & $<$0.001 & 0.069 & *** \\
Aligned & Met.~Acc.& -0.8345 & -0.9940 & $<$0.001 & 0.380 & *** \\
C1.x    & Utility  & 0.9100 & 0.9098 & 0.959 & 0.001 & n.s. \\
C1.x    & Cosine   & 0.6624 & 0.6653 & $<$0.001 & -0.024 & *** \\
C1.x    & Met.~Acc.& -0.8668 & -0.9940 & $<$0.001 & 0.286 & *** \\
C2.x    & Utility  & 0.9115 & 0.9098 & 0.707 & 0.006 & n.s. \\
C2.x    & Cosine   & 0.6761 & 0.6653 & $<$0.001 & 0.086 & *** \\
C2.x    & Met.~Acc.& -0.8746 & -0.9940 & $<$0.001 & 0.275 & *** \\
C3    & Utility  & 0.9179 & 0.9098 & 0.227 & 0.029 & n.s. \\
C3    & Cosine   & 0.6831 & 0.6653 & $<$0.001 & 0.141 & *** \\
C3    & Met.~Acc.& -0.5293 & -0.9940 & $<$0.001 & 1.041 & *** \\
\hline
\end{tabular}
\end{minipage}

\vspace{0.5em}
\caption{Parametric independent-samples $t$-tests on row-level data for both macro (aligned) and micro levels. Equal variance between datasets is assumed. Met.~Acc.\@ = Metric Acceptability, Cosine = Semantic Similarity. $^{***}p < 0.001$; $^{**}p < 0.01$; n.s.\@ $p \geq 0.05$.}
\label{tab:app_param_ttests}
\end{table}

\begin{table*}[htbp]
\centering
\footnotesize
\resizebox{\linewidth}{!}{
    \begin{tabular}{ccp{0.95\linewidth}} 
    \multicolumn{2}{c}{\textbf{Original}} & Anyone else using sugar (table sugar, candy, ...) for OCD?  Seems quite effective for me, but I'm not sure how good it's in longer term.  I speculate that it only gives temporary boost while possibly lowering serotonin production in longer term. \\
    \midrule
    \multirow{3}{*}{\textbf{Config 1.1}} & 250 & I am now a \$100, \$100 and I am now just a little bit of a little money for a little little bit, I think I need to be very much of a bit of money for this little little thing, I am really just a bit for a \\
     & 750 & Mysterious sugar is a good thing, but I don't know if it's beneficial for me, or if it just makes my mood worse. I have a few things I need to try, but it's not that great a thing. I think it's a \\
     & 1500 & I have heard that it works for some people, but I'm not sure how much longer. I think it may be a short term effect. I would like to see it reduced in dosage. I also think it might be beneficial to have some sort of sugar in the \\
    \midrule
    \multirow{3}{*}{\textbf{Config 1.2}} & 250 & I was not aware of this. I was not. I didn't know about it until I was. I did. I looked at it. I wasn't aware of it. It was a bit of a story. I'm not aware. I really did. It \\
     & 750 & I think it's safe to use this stuff, but I'm not sure how much it's worth. I think it might be helpful for short-term memory loss. I'm also not sure if it's helpful in the long term. I would like to see it \\
     & 1500 & I have heard of this type of medication for anxiety, but I'm not sure how effective it is for short term. I think it may be helpful for longer term. But I don't know if it's safe to use it for short-term or long-term \\
    \midrule
    \multirow{3}{*}{\textbf{Config 2.1}} & 250 & In the field of the night, I found a lot of the food in the night and found a little in the morning, but not much in the way of the day. I took it to the doctor for a test, but found nothing in the side of the \\
     & 750 & I have no idea if it's a good idea to use it for short term or long term. I would like to see if it helps with anxiety or if it just keeps increasing my blood sugar levels. I think it's just a matter of time before I feel \\
     & 1500 & I've tried it and it works great. It's just that it's a little bit too much for me, but I think it's worth it. I think its a little too much sugar for me. I like it but it's not really effective in short \\
    \midrule
    \multirow{3}{*}{\textbf{Config 2.2 (R)}} & 250 & I was able to get a bit of a heart attack. I was able for me to be able to do a lot of breathing. I did not feel like I could do it. I had to give up a lot. I found out that I was not able \\
     & 750 & I have tried it for a while now and it works great. I have a headache, but it's not as bad as I thought. I think it's a little high. I can't remember if it's for fun or for a short spell. I'm not \\
     & 1500 & I have tried to get some sort of treatment for anxiety, but it seems to work only for short term. I think it may be a little too much for short-term. I also have a problem with sugar in my food, which can be a bit toxic. \\
    \midrule
    \multirow{3}{*}{\textbf{Config 2.2 (S)}} & 250 & I don't know what to drink. I don't drink much. I drink it for a day. I wonder what happens to my blood. I'm not sure. I think it's a good thing. I find it to be a good way to get to \\
     & 750 & I have tried to use this as a temporary boost to my mood. It seems to work, but I have a problem with it. I think it may be that it's a temporary effect. I have tried a lot of other drugs and it seems to reduce my mood \\
     & 1500 & This is a fairly standard procedure for short term pain, but it may be helpful in the long term. I have read that it may help reduce inflammation in the brain. I would like to try it out for a while. It may also help reduce pain in the short \\
    \midrule
    \multirow{3}{*}{\textbf{Config 2.3 (120)}} & 250 & Regarding the ``says'' and ``I'm not sure if it's a good thing to do, but I'm not really sure if I'm saying it's not a good ``tort, which I'm also not sure I'm sure of, but it's \\
     & 750 & Steroids: I'm not sure what's going on with my head. I'm just trying to get used to it. I don't know what's in it for me. I think it's just a temporary boost to my mood. I can't seem to \\
     & 1500 & Sugar is good for anxiety, but not for depression. It's not really effective in reducing inflammation, but it's a good thing. I think it's safe to use it as a temporary boost to anxiety. I'm not sure how long it will last. I \\
    \midrule
    \multirow{3}{*}{\textbf{Config 2.3 (1800)}} & 250 & In my personal opinion, I would consider it a good idea to go to the bathroom, I could probably go to a more extreme, but I don't want to go with this. I would go with it, I feel it would be better to go for a more \\
     & 750 & It's a bit of a mystery to me, but I think it's worth mentioning. It's a little bit of an anxiety-inducing drug, but it's also a lot of fun to use. I think the only thing I've tried is to use it as \\
     & 1500 & This is a very interesting study, especially since it shows that it can be used for short term or long term benefit. It may also provide a boost to serotonin (lowering anxiety), which is a good thing for short-term. But it may also reduce the effects \\
    \midrule
    \multirow{3}{*}{\textbf{Config 2.3 (720)}} & 250 & * I've been trying to get some of these and I've found that it's actually more likely to be a bit of a risk than just a few of the usual things. I've also been trying it and it's been a bit more than just an occasional case \\
     & 750 & I've been using it for a while now and it seems to work well. I think it's just a short term memory boost. I also have a few other things that I need to try. It's also possible to reduce sugar consumption in the short term. But \\
     & 1500 & It seems to work well for short term, but is it worth it to add more sugar to the diet? Or is it just a placebo? I think so. I've been using it for a while now and it seems to help reduce my anxiety. But I'm \\
    \midrule
    \multirow{3}{*}{\textbf{Config 3.1}} & 250 & Healthy, active, and sensitive food, as well as new, new, and interesting drinks. Also, it's a new category for me, and I'm interested in how much it's added to my food and drink. Also: how much alcohol is added \\
     & 750 & My husband's family used a sugar-based drug for a few weeks, but it's only temporarily suppressed inflammation. I've noticed that it's also causing some inflammation in the brain, which can lead to hyper sensitivity. I'm concerned that it may cause inflammation in some \\
     & 1500 & My doctor prescribed a low-dose version of a stimulant for insomnia, but it only works for short-term use. I'm not sure how much it's actually helping, but I suspect it's a temporary boost to cognitive function. I've also tried a placebo \\
    \midrule \midrule
    \multirow{3}{*}{\textbf{DP-BART (Base)}} & 250 & The ``t-t''s, and the ``res'' (t-s) for the ``t'' and ``c'' (c-t)s'' for the population, the ``c-s'', and their ``res'', ``t'', `` \\
     & 750 & There is no such thing as safe for me, but there is some sort of chemical reaction. What is safe for you, and what is not for me? I don't know. I'm not sure if it's safe for myself, but I think it's \\
     & 1500 & I don't know. Maybe it's a placebo effect. I'm not sure. I think it may be a placebo. I don't really know. It could be a long-term or short-term effect. It might be a medium-term one. \\
    \midrule
    \multirow{3}{*}{\textbf{DP-BART (Large)}} & 250 & Oent be be be being being be being be be, be beed be be in be being, being being being, be being in be be down be be for be be it being being in being be, being be in being being for be being down be \\
     & 750 & In the past, I have been able to do this in the form of a self-serum. I have done this for the past 12 months. I can see that it's good for me. In fact, I've been doing it for the \\
     & 1500 & I'm not sure what to do about this. I've been thinking about it for a while. I'm thinking about using it as an anti-depressant. I have been using it for several years. It seems to be working for me. I'm \\
    \bottomrule
    \end{tabular}    
}
\caption{Privatization examples from the Reddit Mental Health dataset.}
\label{tab:examples_reddit_mental_health}
\end{table*}

\begin{table*}[htbp]
\centering
\footnotesize
\resizebox{\linewidth}{!}{
    \begin{tabular}{ccp{0.95\linewidth}} 
    \toprule
\multicolumn{2}{c}{\textbf{Original}} & Never had a Glitch.: I have ordered a few things now including used equipment and everything has come on time and as described. I also have had very helpful phone conversations too. always a pleasure. \\
\midrule
\multirow{3}{*}{\textbf{Config 1.1}} & 250 & I have been very very very pleased and I have been able to give my orders to the people I have bought and I am very, very, I have had very good time. I have not had \\
 & 750 & Satisfactory. I have ordered a few items for myself and it has been very helpful. I am very pleased with the quality and the service. I can now order a few things for myself. I \\
 & 1500 & I have had a few problems with my phone but have been able to get a few things done. I have had very little trouble with the software I have used. I am now able to send a few \\
\midrule
\multirow{3}{*}{\textbf{Config 1.2}} & 250 & I have been in the process of getting the information to the right people and I have been on the phone for a number of times. I have not had the information as I have to. I am \\
 & 750 & The following is a list of items I have ordered. I have used a few items that I have liked and have enjoyed the experience. I am very pleased with the results. I had a great time and \\
 & 1500 & I have had a few problems with the internet. I have had to order a few items and have had very little response. I am very happy with the results. I also have had some very helpful phone \\
\midrule
\multirow{3}{*}{\textbf{Config 2.1}} & 250 & I have a great deal of a gratitude for this. I have a good understanding of this. It is a great way to get a job and I am a great help. I am not a bad \\
 & 750 & I have a few questions for you. I have ordered a few items and am very happy to see that everything is going great. I also have a lot of questions about the shipping costs and how to \\
 & 1500 & I have a few things to look forward to. I have ordered a few items and have been very happy to receive them. I also have a list of all the required equipment and services. I will \\
\midrule
\multirow{3}{*}{\textbf{Config 2.2 (R)}} & 250 & I have a chance to meet you in the chance to see you again. I am now able to give you a chance. I have a good chance to find out if I can find a job in \\
 & 750 & I have had a great experience with this. I have ordered a few items and have been very happy with them. I would like to order a few more items and will be happy to get them. \\
 & 1500 & I have a great deal to offer. I have ordered a few things now and it is very nice to have them delivered. I also have a lot of great reviews and I have had a great time with \\
\midrule
\multirow{3}{*}{\textbf{Config 2.2 (S)}} & 250 & The first order is for the new order and the first order has been ordered. I have been told that it is a little more complicated but it is also a good way. I am now in the \\
 & 750 & My phone has been working and I have had a lot of questions and answers. I have been able to get a lot done and I am very pleased with the results. I am now able to order a \\
 & 1500 & I have a few things to review. I have ordered a few items and have been very pleased with the quality. I also have a couple things to try and get ready for a trade. I am very \\
\midrule
\multirow{3}{*}{\textbf{Config 2.3 (120)}} & 250 & I have been able to get my money and I have been very happy. I have a few of my own and I'm still able to pay for my money. I'm very happy to have my money \\
 & 750 & All in all a great experience. I have had a great time and have had many good reviews. I am very happy with the results and have been able to get a few things ordered. I also have \\
 & 1500 & I have had a few problems with my computer but have been able to get a few things done. I have been very helpful and have been to a few places and have had very good results. I am \\
\midrule
\multirow{3}{*}{\textbf{Config 2.3 (1800)}} & 250 & The company has been very helpful and has helped me to have a clear and clear and have been able to provide the necessary information. I have been very happy with the product and have had a great amount of \\
 & 750 & I have had a few problems with the shipping. I have had to cancel all orders and it has come to an end. I am very pleased with the results. I will continue to send samples to the \\
 & 1500 & All in all, a very good experience. I have ordered a few things now and have been very pleased with the results. I also have had a few phone calls and emails from the suppliers. I am \\
\midrule
\multirow{3}{*}{\textbf{Config 2.3 (720)}} & 250 & I have been in the market for a while. I have been very happy with the results. I think I will be in the new position. I am very happy. I can do it. I hope \\
 & 750 & I have had a few good reviews. I have used a few things to help me out. I just have to say thank you very much for everything. I hope to have a few more to add soon \\
 & 1500 & Lately I have had a few calls and emails from customers. I have been very helpful and have had many great experiences. I just have to say thank you very much for all the support. I hope \\
\midrule
\multirow{3}{*}{\textbf{Config 3.1}} & 250 & Siri tip: I'm working on a new website and online reservation reservation. It's a great upgrade from the original site and a great online experience. I also made a new online reservation for a \\
 & 750 & My wife and I have ordered several items over the years, including a few new ones. I have had great reviews from reputable retailers and have received great reviews. I also have had my own online ordering process \\
 & 1500 & My husband and I have used a few computers over the years. I have ordered a few items online and have received great reviews. I also have used several online ordering services, including a free online ordering service \\
\midrule \midrule
\multirow{3}{*}{\textbf{DP-BART (Base)}} & 250 & And now, I have a lot of time and money to do all of the things I want to do, and I can’t quite make it all of it, but I have all of \\
 & 750 & Thank you very much. We have had a great time. We are very, very happy to have had so many wonderful people. I have had many, very nice people. Thank you so much. \\
 & 1500 & All in all, I have had a great time. I have a few things to look forward to. I am very happy with the results I have received. I do not have a lot of time \\
\midrule
\multirow{3}{*}{\textbf{DP-BART (Large)}} & 250 & ( in, in,,/,,,'s,, in/,/'s,///,'s//(,, Queen,/(/, in in,/ in \\
 & 750 & I have been using this service for a few years now. I have ordered a few things from this company and have been pleased with the results. I also have had very good service. I \\
 & 1500 & Never had a problem with any of the products. I have ordered a few things and everything has been perfect. I also have had a few problems with the product.Never had any problems with anything. \\
\bottomrule
\end{tabular}
}
\caption{Privatization examples from the Trustpilot dataset.}
\label{tab:examples_trustpilot}
\end{table*}

\begin{table*}[htbp]
\centering
\footnotesize
\resizebox{\linewidth}{!}{
    \begin{tabular}{ccp{0.95\linewidth}} 
    \toprule
\multicolumn{2}{c}{\textbf{Original}} & It's a chain and it has what I need.  It's reasonably convenient. Not yuppified at all.  Service is minimal but I am usually able to find the dog-related supplies I need right away. \\
\midrule
\multirow{3}{*}{\textbf{Config 1.1}} & 250 & It's not a bad thing to have a mous of the m, but it is not a thing to do, and I am not sure I am doing anything to help, but I am, and it is a thing \\
 & 750 & It's a good thing I have it. I need to get the supplies I need. I'm not a big fan of the store. I have to find the supplies that I need when I need them. I am not a \\
 & 1500 & It's a good thing I have it. It's a reasonably priced place to store my supplies. It has a decent supply. I am not sure I need to use it for the necessities. I have to find the right one \\
\midrule
\multirow{3}{*}{\textbf{Config 1.2}} & 250 & I'm sure I'm looking at a new place to look at the road and I'm sure that I'm getting a good look at a road and a look at what I was looking at. I'm also sure I was \\
 & 750 & It's a good thing I have it. I like it. It's convenient. I can get around the city. I have a lot of stuff I need to get done. I am not a big fan of the store- \\
 & 1500 & It's a good thing I have it. I need to find the supplies I need. I have a lot of stuff I need, but I am able to get the basics right away. I am also able to find a place \\
\midrule
\multirow{3}{*}{\textbf{Config 2.1}} & 250 & It's a great way to get the food and it's great. It's a good way to work out. It also helps to keep the food on the table. It helps to have a good food and a great food \\
 & 750 & It's a good thing. It's a lot of work. I need to find the supplies I need. It takes time. I have a lot to get everything I need, but I am able to get the supplies. \\
 & 1500 & It's a simple thing to do. I have a backpack full of supplies. I can carry everything I need. It's a good size. I am able to find the supplies I need right away. I also have a \\
\midrule
\multirow{3}{*}{\textbf{Config 2.2 (R)}} & 250 & It is not just the food. I am not sure I want to eat. I want the food to be free. I will be able to eat the food I need. I can't eat the way I want. I \\
 & 750 & It's a nice place and I like it. I like to shop. It's a bit pricey but I like my stuff. I have a lot of stuff I need. I can get it all done. I am not \\
 & 1500 & It's a nice place and it has a lot of storage. I need to find the supplies I need. It's a little pricey but I am able to get the supplies that I need right away. I have a lot to \\
\midrule
\multirow{3}{*}{\textbf{Config 2.2 (S)}} & 250 & It's a good thing I was able to get it. I was going to be able to see the money. I had to get a good deal. I could have been able to pay for it. It was a lot \\
 & 750 & It's a simple thing to do. I have a car and I need to get the supplies I need. I need a bike. I can get the basics I need right away. I am usually able to find the right one \\
 & 1500 & It's a small shop and it has a lot of supplies. I need to find the right one. It's a good place to get the supplies I need. I usually find the cheapest I need right away. It is a \\
\midrule
\multirow{3}{*}{\textbf{Config 2.3 (120)}} & 250 & I'm a good, good, and I'm a little bit on the budget. I'm not a good person. I can get my money on the things I need to get my things on the way. I am also a \\
 & 750 & It's a good place to start. It's convenient. It has a lot of storage space. It can be used for food and other things. It is also convenient to get the supplies I need. I have a small office \\
 & 1500 & It's a small house and it's reasonably priced. It's a good size. I have a lot of stuff I need. I'm not sure what I need but I'm usually able to find the right place to put my \\
\midrule
\multirow{3}{*}{\textbf{Config 2.3 (1800)}} & 250 & Rest place is a good place to get the stuff I need. I know you have a place to go. I think I have a good choice. I have some stuff to get. I feel comfortable. I am sure you know \\
 & 750 & It's a convenience store and it has what I need. I have a cart that I need to get the supplies I need for the day-to-day. It's a good place to get supplies that are needed. I \\
 & 1500 & It's a good thing I have it. I need the supplies I need. It's a small place and it's easy to get the supplies you need. I have the right equipment. I can usually find the supplies and I \\
\midrule
\multirow{3}{*}{\textbf{Config 2.3 (720)}} & 250 & It's been my privilege to have my own place of my own, and it's been so nice to be able to have it done. It's been a few times that I've been able to get my own food and it \\
 & 750 & It's a good place to get supplies and I'm able to get the job done quickly. I'm not a big fan of the store but I'm usually able to find the right place to go for the essentials I need. \\
 & 1500 & It's a simple thing to do. I have a few things I need. I can find the right place at a reasonable price. I also have a fairly reliable supply of water. I am usually able to find the nearest water \\
\midrule
\multirow{3}{*}{\textbf{Config 3.1}} & 250 & It's a weekly weekly weekly subscription. It's easy to buy and check out. It also comes with a weekly subscription subscription. Each week, it's available for a monthly subscription. Every week, I receive a weekly shipment \\
 & 750 & It's a small office and it has all the necessary supplies. I'm reasonably priced, but I don't have the full-priced office space I need. I can usually find the nearest one online, and I usually find a \\
 & 1500 & It's a small office and it has a small supply of supplies. I'm usually able to find the supplies I need quickly, but I'm not always able to get the exact item I need. My supply chain is small, \\
\midrule \midrule
\multirow{3}{*}{\textbf{DP-BART (Base)}} & 250 & It is a pain in the mime and I can do a lot of things. I am more and I have more and more money. I can go on and on and I am no longer able to find the money. \\
 & 750 & It's a good place to start. I have a lot of stuff I need. I need a place to store my stuff. I am not a very good person. I don't know what I'm dealing with. I \\
 & 1500 & It's a good thing. It's a very convenient place to get the supplies I need. Very convenient. And it's very convenient. I can get the things I need at the same time. \\
\midrule
\multirow{3}{*}{\textbf{DP-BART (Large)}} & 250 & In [ [ [ Met's [ [ National [ [ D's [ National National National [ Met [ [ X's Met's National National Met's Met [ Met Met's One [ National's National [ National Met [ \\
 & 750 & It's a little bit of a mystery, but it's a good one. It's a small town in the middle of a town.It's not a big town, but I have a good idea of what \\
 & 1500 & It's a great store. It's a good location.It's also a good store.It is a great place.It has a great selection.It also has a good selection.I love this store.I'm \\
\bottomrule
\end{tabular}
}
\caption{Privatization examples from the Yelp dataset.}
\label{tab:examples_yelp}
\end{table*}

\end{document}